\documentclass[pmlr,twocolumn,10pt]{jmlr}

\usepackage{booktabs}
\usepackage{amsmath}
\usepackage{graphicx}
\usepackage{multirow}
\usepackage{colortbl}
\usepackage{pgfplots}
\pgfplotsset{compat=1.18}

\definecolor{gain}{HTML}{2C875A}
\definecolor{loss}{HTML}{BC4238}

\jmlrvolume{}
\jmlryear{2026}
\jmlrworkshop{Machine Learning for Health (ML4H) 2026}

\title[Which Model, For Which Variable?]{Which Model, For Which Variable?\\
  Per-Variable Forecast Combination in Sparse ICU Time Series}

\author{\Name{Marius Fracarolli} \Email{fracarolli@cl.uni-heidelberg.de}\\
 \addr Department of Computational Linguistics, Heidelberg University, Germany\\
 \addr HIDSS4Health --- Helmholtz Information \& Data Science School for Health,
 Karlsruhe/Heidelberg, Germany
 \AND
 \Name{Stefan Riezler} \Email{riezler@cl.uni-heidelberg.de}\\
 \addr Department of Computational Linguistics \& Interdisciplinary Center for
 Scientific Computing (IWR), Heidelberg University, Germany}

\begin{document}
\maketitle

\begin{abstract}
Clinical forecasting papers typically report a single best architecture. We show that on multivariate ICU
physiology this question conceals variable-specific differences: no architecture is best on every variable, and \emph{which} model wins changes
from variable to variable. Across five ICU databases (MIMIC-III, MIMIC-IV, eICU, HiRID, and \textsc{UMM} (University Medicine Mannheim), a German
single-center ICU) we forecast 94 channels over a 24-hour horizon with a pool of 14
models, and combine them with weights fitted \emph{per variable}. On test, the combination beats the best single model on all five databases by
6.1--11.0\,\% mean squared error (mean 8.7\,\%). The gain is concentrated where clinical data is
thinnest: the sparsest third of channels improves by 11.9\,\% and laboratory values by 7.5\,\%,
against 3.1\,\% for the densest third. Rank-based harmonic weighting of the best five models
per variable, with no fitted weights at all, recovers about four-fifths of the gain.
Extending the pool afterwards with a per-channel gradient-boosted tree, ARIMA and ETS makes the
tree the validation-selected best single model on all five databases; per-variable combination
still improves on it by 2.9--5.9\,\% (significant on four of five), while equal weighting of the
same members does not.  As on univariate benchmarks, learned combination beats
simple averaging --- but only once the weights are constrained. We also report which model to prefer when, and show that the ordering
changes with database size.
\end{abstract}

\paragraph*{Data and Code Availability}
All code for harmonization, training, evaluation and combination is publicly available at \url{https://github.com/StatNLP/ML4H_2026_Per-Variable-Forecast-Combination}. The four public databases (MIMIC-III, MIMIC-IV, eICU and HiRID) are available via PhysioNet under their respective data use agreements; \textsc{UMM} is a single-center clinical database that cannot be shared.

\paragraph*{Institutional Review Board (IRB)}
While IRB approval is not required for the scientific usage of the four public databases, retrospective analysis of the \textsc{UMM} data was approved by the Medical Ethics Commission II of the Medical Faculty Mannheim, Heidelberg University, Germany, with waiver of patient consent (reference numbers 2016-800R-MA, 2016-840R-MA and 2023-851).

\section{Introduction}

A clinical time-series forecasting paper typically ends with a table in which one architecture has
the lowest error, and that architecture is described as the best model for the task. This paper
argues that on multivariate ICU physiology the single-best-model framing is suboptimal. The target is not one series but
a 94-channel state vector mixing daily laboratory values, continuously sampled vital signs,
medications that are administered, not measured, and derived scores. The observation process
is extremely sparse --- 8.8--18.6\,\% of (variable, hour) cells carry a measurement --- and the
sparsity is itself informative, because sicker patients are sampled more often. Furthermore, the channels are
heterogeneous in a way univariate benchmarks are not: a creatinine trajectory and a noradrenaline
infusion rate have almost nothing in common as forecasting problems. Asking which
\emph{architecture} is best therefore assumes one inductive bias dominates the whole panel. We show that fitting combination weights separately for each variable --- a per-variable
stacked generalization --- recovers a consistent 6.1--11.0\,\% error reduction over the best single
model on every database we examine, on a test set spent exactly once.

\paragraph{Contributions.}
\begin{enumerate}
\item \textbf{Per-variable forecast combination for sparse clinical panels.} We formulate stacking
with variable-specific weights, shared across windows and forecast horizons rather than tied to a series, and give an efficient estimator that makes the full combiner ladder a seven-minute CPU job over
cached predictions. 
\item \textbf{A five-database held-out confirmation.} Per-variable forecast combination improves over the best single model on all five databases
(6.1--11.0\,\%), with the combiner chosen on validation before test was touched.
\item \textbf{Sparse channels gain most.} Improvement tracks channel sparsity: averaged over the five
databases the sparsest third of channels gains 11.9\,\% against 3.1\,\% for the densest. We also
show that the metric conventionally used for such panels is dominated by near-unobserved channels,
and report it with a minimum-support filter.
\item \textbf{Guidance on which model to use when}, from per-channel win shares: no model is
uniformly preferable, and the leader changes between databases. A dense autoregressive
decoder is the most consistent choice at any size; the set-function encoder STraTS leads on HiRID
and MIMIC-IV and wins fewest channels on the smallest and densest database, where an
architecture built for irregular sampling has least to exploit; linear models are competitive only on the small
databases; and the zero-shot foundation models are included for their decorrelation and not their accuracy.
Adding a gradient-boosted tree changes the picture: it wins most cells on every database
(Section~\ref{sec:lgbm}), and per-variable combination still improves on it.
\end{enumerate}

\section{Related Work}\label{sec:related}

\paragraph{Forecast combination.} That combinations of forecasts outperform their members is one of
the oldest results in the field \citep{BatesGranger:69}, and the persistent finding that
\emph{simple} combinations beat \emph{estimated} ones --- the forecast combination puzzle --- has
survived decades of competition evidence \citep{MakridakisETAL:20,MakridakisETAL:22}. Stacked
generalization \citep{Wolpert:92} and stacked regressions \citep{Breiman:96stacked} established the
modern framing, with the key practical requirement that the combiner be fitted on out-of-fold
predictions; greedy ensemble selection \citep{CaruanaETAL:04} is the constrained alternative we also
evaluate. The ambiguity decomposition \citep{KroghVedelsby:95} makes the mechanism exact. For a weighted
average $\bar y = \sum_m w_m \hat y^{(m)}$ with $\sum_m w_m = 1$, under squared error,
\[(\bar y - y)^2 = \textstyle\sum_m w_m (\hat y^{(m)}-y)^2 - \sum_m w_m (\hat y^{(m)}-\bar y)^2,\]
an identity, not a bound: ensemble error is mean member error minus the \emph{ambiguity}, the
weighted variance of the members about their own mean. Combination therefore pays exactly in proportion to member disagreement, and we report the
ambiguity term directly (Figure~\ref{fig:maps}). \citet{PerroneCooper:92} take the further step
of choosing weights from the members' error \emph{covariance} and not their individual errors,
showing that the resulting combination beats every member --- the argument our per-channel Gram
matrix implements. \citet{HansenSalamon:90} give the original variance-reduction result and
\citet{GanaieETAL:21} survey the deep-learning instantiation. Closest to us in setting,
\citet{SchamoniETAL:22} ensemble neural networks for early sepsis prediction in the ICU, so
disagreement-weighted combination is already validated on clinical time series; we extend it from a
single target to a 94-channel panel with per-channel weights.

\paragraph{Stacking for time series.} Recent work \citep{BoschETAL:25} has systematized stacking for
forecasting at benchmark scale, establishing out-of-fold protocol requirements and comparing
combiner families.
Their conclusion that learned stacking beats both individual models and simple aggregation
agrees with ours. Their evidence base is 50 \emph{univariate}, general-purpose datasets
\citep{AnsariETAL:24:Chronos,WooETAL:24:Moirai} with no clinical application, filtered so that
enough data exists to train base and stacker models, and their unit of prediction is the series
itself, so there is no feature axis along which weights could be tied. Our channels are 94 per
window, 8.8--18.6\,\% observed, and neither independent nor comparably supported. Whether stacking
survives clinical sparsity is therefore open. Section~\ref{sec:findings} shows that it does.

\paragraph{Clinical time-series forecasting.} Irregular sampling has motivated architectures built
around observation triplets instead of dense grids \citep{TipirneniReddy:21}, and the
cross-database reproduction of \citet{StaniekETAL:24mlhc} sets our pool and its 4-hourly,
24-hour-horizon protocol. Zero-shot foundation models
\citep{WooETAL:24:Moirai,PodestETAL:26:TiRex2} are a natural addition and we include
two; persistence remains the incumbent baseline \citep{HaugsdalETAL:23:Persistence,HewamalageETAL:23}.

\section{Data and Setup}

\subsection{Databases}

We use five ICU databases: MIMIC-III \citep{JohnsonETAL:16}, MIMIC-IV \citep{JohnsonETAL:23},
eICU \citep{PollardETAL:18}, HiRID \citep{HylandETAL:20}, and \textsc{UMM} (University Medicine Mannheim, Heidelberg University), a German single-center ICU whose data are not publicly available.
All five are harmonized onto a shared 94-channel schema and gridded to at most one entry per
(variable, hour) cell, with unrecorded cells marked unobserved in the mask; windows carry a 24-hour
input and a 24-hour horizon and are drawn from the first seven ICU days at a 4-hour stride. Fixed at one seed, splits are 64/16/20 (train/validation/test) by patient (by admission for HiRID, where readmissions are not linkable), so no stay appears in two splits.

\begin{table}[t] 
\centering\small\setlength{\tabcolsep}{2.2pt}
\caption{Databases after harmonization. \emph{Stays} and \emph{windows} are totals over all splits;
the 64/16/20 breakdown is in Appendix~\ref{app:data}. \emph{Density} is the fraction of
(variable, hour) target cells carrying an observation; \emph{charted} is how many of the 94 channels
the database ever records. \emph{Patients} are unique individuals; HiRID releases one record per
admission with no cross-admission linkage, so its stays and patients coincide by construction.}
\label{tab:data}
\begin{tabular}{lrrrrr}
\toprule
 & stays & patients & windows & density & charted \\
\midrule
MIMIC-III & 28\,514 & 22\,338 & 431\,111 & 11.0\% & 85 \\
MIMIC-IV & 46\,220 & 35\,706 & 693\,548 & 11.4\% & 89 \\
eICU & 77\,115 & 63\,859 & 1\,065\,028 & 8.8\% & 79 \\
HiRID & 8\,621 & 8\,621 & 136\,683 & 13.3\% & 66 \\
\textsc{UMM} & 4\,490 & 4\,228 & 97\,061 & 18.6\% & 80 \\
\bottomrule
\end{tabular}
\end{table}

Table~\ref{tab:data} makes the central difficulty explicit. Even the densest database leaves four
of every five target cells unobserved, and the databases differ a lot in size: eICU has 17.2 times
\textsc{UMM}'s stays. Sparsity
determines where combination pays, and size determines whether the combiner can be fitted at all.

\subsection{Evaluation metric}

Write $\hat y_{nhf}$ for the prediction of window $n \in \{1..N\}$ at horizon step
$h \in \{1..24\}$ for channel $f \in \{1..94\}$, $y_{nhf}$ for the observed value, and
$m_{nhf} \in \{0,1\}$ for the mask, which is 1 when that cell carries an observation.
Let $C$ be the set of channels the database ever charts and $|C|$ its size. All losses are masked
to observed cells, and we report the per-channel mean
\[
\mathrm{MSE}_{\text{per-var}} = \frac{1}{|C|}\sum_{f \in C}
  \frac{\sum_{n,h} m_{nhf}\,(\hat y_{nhf}-y_{nhf})^2}
       {\sum_{n,h} m_{nhf}},
\]
with $n$ running over windows and $h$ over the 24 horizon steps. On test, error grows 1.2--1.9$\times$ from $h{=}1$ to $h{=}24$ while the combination gain stays positive but falls by half or more (Appendix~\ref{app:horizon}).
Each channel contributes its own mean squared error, and the channels are then averaged with equal
weight. Values are z-scored per channel using training-split statistics, so an error of 1.0 is one
training standard deviation of that channel and the numbers are comparable across channels.

We use squared error because the ambiguity decomposition that motivates combination
(Section~\ref{sec:related}) is an identity only under squared error; it does not hold for MAE (mean absolute error), and
probabilistic scores such as CRPS (continuous ranked probability score) would require the members to emit distributions, which several of
ours do not. Weighting channels, not observations, suits a
clinical panel --- a rare but decision-relevant assay should not be drowned out by a continuously monitored
vital. However, channels with a
handful of observations can dominate a group mean.  One HiRID medication channel with twelve observed cells supplies 41 \% of its group's mean. We therefore apply a minimum-support
filter of 100 observations to group-level tables (Appendix~\ref{app:win}).

\subsection{Model pool}\label{sec:pool}

The pool holds 14 models per database (full listing, Appendix~\ref{app:models}): two trivial
baselines (\textsc{null}, all-zeros in normalized space, and \textsc{locf},
last-observation-carried-forward); a linear model, DLinear \citep{ZengETAL:23}; a direct
multi-step dense model, Dense DMS \citep{StaniekETAL:24mlhc}; two recurrent models
(GRU \citep{ChoETAL:14:GRU}, LSTM \citep{HochreiterSchmidhuber:97}); a state-space model, Mamba \citep{GuDao:23}; a patch transformer, PatchTST
\citep{NieETAL:23}; two long-horizon transformers, Informer \citep{ZhouETAL:21} and Autoformer
\citep{WuETAL:21}; an irregular-sampling clinical transformer, STraTS \citep{TipirneniReddy:21}; an
autoregressive dense decoder, DenseAR \citep{StaniekETAL:24mlhc}; and two zero-shot foundation
models, Moirai \citep{WooETAL:24:Moirai} and TiRex-2 \citep{PodestETAL:26:TiRex2}. Every learned member is
tuned per database rather than sharing one configuration, and pool membership is decided on
validation error under a single rule applied to every family; the selected configurations are listed
in Appendix~\ref{app:models}.

Because the foundation models' pretraining corpora are partially undisclosed, we cannot rule out prior exposure to our four public databases. We therefore do not interpret their absolute performance, including them solely for ensemble diversity as the pool's least correlated learned members (Appendix~\ref{app:div}).

Training uses one fixed protocol for every model, schedule and database: early stopping on
validation with patience 6, and learning-rate reduction on plateau with patience 3. 

Even LOCF, one of the trivial baselines, has its place because it wins 2.5--3.4\,\% of well-supported test cells outright
(Appendix~\ref{app:win}).

Base predictions are cached to disk once, which reduces the entire
combiner ladder over five databases to a seven-minute CPU job, and no combiner ever re-runs a network.

\section{Method}

\subsection{Per-variable stacking}

Let $\hat y^{(m)}_{nhf}$ be model $m$'s prediction for window $n$, horizon step $h$, channel $f$,
with observation mask $m_{nhf}$. We seek weights $w_{fm}$ --- tied across windows and (in our main
variant) across horizon steps, but \emph{free across channels} --- minimizing masked squared error.

The objective is a masked sum of squared errors, $\mathrm{SSE}$, and because it is quadratic in $w$
it does not depend on the predictions individually, only on two summaries of them. Accumulating
those summaries once lets every combiner below be fitted and scored without ever revisiting a
prediction. The summaries
are a per-channel Gram matrix $G_f$, holding how strongly each pair of models $(m,m')$ agrees on
channel $f$, and a cross-term $b_f$, holding how strongly each model agrees with the truth:
\begin{align*}
G_{f m m'} &= \textstyle\sum_{n,h} m_{nhf}\,\hat y^{(m)}_{nhf}\hat y^{(m')}_{nhf},\\
b_{fm} &= \textstyle\sum_{n,h} m_{nhf}\,\hat y^{(m)}_{nhf} y_{nhf},
\end{align*}
$G_f$ is the piece that heuristics lack: it records how strongly each \emph{pair} of members agrees,
so the learner can discount two models that make the same mistake and reward one that captures
variance nobody else does --- which is why the zero-shot foundation models earn weight despite
middling accuracy. Ranking-based rules such as \textsc{harm-}$k$ use each member's error on its own and never the cross-model products.
Then the error of \emph{any} weight vector follows in closed form,
$\mathrm{SSE}_f(w) = w^\top G_f w - 2 w^\top b_f + \sum_{n,h} m_{nhf} y_{nhf}^2$, so a combiner is
fitted by minimizing a quadratic in $M=14$ unknowns per channel, one weight per pool member, rather than by another pass over
data. The last term does not depend on $w$ and only fixes the offset.  Both statistics are plain sums over windows, which also makes them additive: partial sums over
disjoint groups of stays can be recombined by addition, and that is what makes the bootstrap and the
randomization test in Appendix~\ref{app:unc} affordable.

\subsection{Combiners}

We evaluate a ladder of increasing flexibility (formal definitions, Appendix~\ref{app:comb}):
\textsc{best-single}, the validation-best member, which every percentage below is measured against; per-variable \textsc{select},
which uses only the best member on each channel; unweighted \textsc{mean} and \textsc{g-equal-3};
inverse-MSE weighting; rank-based harmonic weighting over the top $k$ members
($w \propto 1, \tfrac12, \dots, \tfrac1k$) \textsc{harm-$k$} following \cite{CormackETAL:09:ReciprocalRank} unsmoothed and on models rather than ranked lists on retrieved items;
per-variable \textsc{ridge}, the unconstrained
least-squares solution $w_f = (G_f + \lambda I)^{-1} b_f$, free to assign negative weights;
simplex-constrained least squares, the same objective restricted to $w \ge 0$, $\sum_m w_m = 1$, so
the result is a weighted \emph{average} whose error on any single channel is bounded by that channel's worst member (projection
following \citealp{DuchiETAL:08}); \textsc{greedy} forward selection with replacement
\citep{CaruanaETAL:04}; and a shrunk simplex variant that interpolates each channel's
fitted solution toward the harmonic ranking prior with strength $n_f/(n_f+\kappa)$: a dense channel
keeps almost all of its fitted weights, while a channel with a handful of observations falls back to
\textsc{harm-5} on that channel, and a channel with no support at all falls back to a single
hour-pooled weight vector estimated from every channel at once. \textbf{The shrunk simplex variant, at $\kappa=500$, is the combiner we fixed on validation before opening test} and is what \textsc{ensemble} denotes throughout.
Shrinking toward the ranking prior rather than toward uniform weights is deliberate: the prior is
already near-optimal (Section~\ref{sec:fixed}), so the fallback costs little where it engages. \textsc{select} and \textsc{ridge} also have hour-resolved counterparts (\textsc{sel-}$h$, ridge-$h$), in which $w$ varies with $h$.

\paragraph{How each combiner is fitted.} None of the combiners needs backpropagation or a training loop. \textsc{ridge} has the closed form $w_f = (G_f + \lambda I)^{-1} b_f$ with $\lambda = 1$ on the unnormalized statistics. The regularizer is not optional: $G_f$ is a Gram matrix of highly collinear predictors (mean pairwise residual correlation 0.82--0.87), and on a thin channel it is rank-deficient outright, with fewer observed cells than members. Adding $\lambda I$ makes it positive definite and the inverse well posed \citep{HoerlKennard:70:RidgeRegression}. 
The simplex-constrained
variants, \textsc{ensemble} among them, are convex programs, so projected gradient descent on the
quadratic above reaches a global optimum; collinear members can leave it non-unique, which is a
reason not to over-read an individual weight.
All of them read only $G_f$ and $b_f$, so the entire ladder costs one pass over the cached
statistics.

\paragraph{What a fitted combination looks like.} The weights are interpretable, and they move with
the character of the channel. On eICU, heart rate --- observed in 18.7 of 24 hours --- spreads across neural members and
gives no weight to copying the last value; its five largest weights are $0.27$ DenseAR, $0.18$ Informer, $0.13$ STraTS, $0.12$
LSTM, $0.12$ GRU. Lactate, 150 times sparser, puts LOCF first outright ($0.29$ LOCF, $0.24$ STraTS,
$0.16$ DenseAR), and noradrenaline --- a treatment decision rather than a physiological signal ---
splits between copying forward and predicting nothing ($0.26$ LOCF, $0.25$ STraTS, $0.21$ Dense DMS,
$0.09$ null). The combiner is not only reweighting networks, but deciding, per channel, how much of
the forecast should be a forecast at all.

\paragraph{Protocol.} Combiner weights are never fitted on data used to score them: during
development on a stay-grouped half of validation, scored on the disjoint half; for the final
confirmation, run once, on all of validation and scored on all of test. Test was never used to
select a combiner, a pool member or a hyperparameter (Appendix~\ref{app:models}).

\paragraph{What ``per variable'' does and does not mean.} The base models are multivariate in
the input and joint in the output. Each receives the full 94-channel window with its
observation mask and emits all 94 channels, so channels inform one another inside a model.
``Per variable'' refers only to the combination layer, which fits one weight vector per channel
over the pooled member forecasts. The loss is masked, so a channel contributes only where a
value was recorded, and within a (variable, hour) cell the target is the last recorded value in
the bin. Feeding the iterative decoders the combined forecast during generation is no deployable
alternative: it helps only when injected at the cells where the target will be observed, which
leaks the future observation mask 
(Appendix~\ref{app:neg}).

\section{Results}
\label{sec:results}

\subsection{Main result}

\begin{table}[t]
\centering\small\setlength{\tabcolsep}{3.5pt}
\caption{Held-out test error (MSE-per-var, lower is better). \textsc{best-single} is the
validation-chosen best member; \textsc{select} commits to the best member \emph{per channel};
\textsc{harm-5} is the rank-based harmonic blend of the best five per channel; \textsc{ensemble} is the
combiner chosen on validation. Weights fitted on all of validation, scored on all of test, spent
once. Percentages are against \textsc{best-single}, which is DenseAR on all five databases. All
charted channels, without the support filter of the group-level tables (Appendix~\ref{app:system}).}
\label{tab:final}
\begin{tabular}{lrrrr}
\toprule
 & \textsc{best} & \textsc{select} & \textsc{harm-5} & \textsc{ens.} \\
\midrule
\textsc{UMM} & 0.6036 & 0.5740 & 0.5505 & \textbf{0.5375} \\
 & --- & $-4.9\%$ & $-8.8\%$ & $\mathbf{-11.0\%}$ \\
HiRID & 0.4865 & 0.4509 & 0.4450 & \textbf{0.4378} \\
 & --- & $-7.3\%$ & $-8.5\%$ & $\mathbf{-10.0\%}$ \\
MIMIC-III & 0.4638 & 0.4460 & 0.4244 & \textbf{0.4184} \\
 & --- & $-3.8\%$ & $-8.5\%$ & $\mathbf{-9.8\%}$ \\
MIMIC-IV & 0.4431 & 0.4366 & 0.4211 & \textbf{0.4143} \\
 & --- & $-1.5\%$ & $-5.0\%$ & $\mathbf{-6.5\%}$ \\
eICU & 0.4157 & 0.4094 & 0.3949 & \textbf{0.3904} \\
 & --- & $-1.5\%$ & $-5.0\%$ & $\mathbf{-6.1\%}$ \\
\midrule
mean & 0.4825 & 0.4634 & 0.4472 & \textbf{0.4397} \\
 & --- & $-3.8\%$ & $-7.2\%$ & $\mathbf{-8.7\%}$ \\
\bottomrule
\end{tabular}
\end{table}

As detailed in Table~\ref{tab:final}, per-variable combination beats the best single model on all five
databases, by 6.1--11.0\,\% (8.7\,\% mean), on held-out data, with the combiner fixed in advance. Committing to
the best model \emph{per channel} also beats the best
single model, but recovers under half as much ($-3.8\,\%$ mean), and a rank-based harmonic blend with no
fitted weights recovers most of the rest ($-7.2\,\%$). So the gain comes from blending several members rather than committing to one, and most of it is
available without fitting anything. Every
number in the last two columns is significant at $p < 10^{-4}$ under a paired randomization test
over stays (Appendix~\ref{app:unc}); the only combiner anywhere that fails significance is the
unweighted mean, on MIMIC-IV. The full ladder is in Appendix~\ref{app:ladder}.

\subsection{Further findings}
\label{sec:findings}

\paragraph{1. Per-channel selection does the work; the weight profile adds little.}
\label{sec:fixed}
Two things happen when the best five members per channel are blended harmonically: the members are
chosen per channel, and they are weighted unequally. Crossing the two isolates them. Taking the five
best members \emph{pool-wide} and weighting them equally recovers $-4.89$\,\% (G-EQUAL-5); choosing the five
best \emph{per channel} and still weighting them equally recovers $-6.97$\,\% (EQUAL-5). Membership is
therefore worth about $2.1$\,pp. Applying the harmonic profile to the per-channel members adds a
further $0.18$\,pp ($-7.15$\,\%, \textsc{harm-5}), while applying it to the pool-wide members \emph{costs}
$0.56$\,pp (G-HARM-5). The profile pays only when rank~1 is that channel's best model, not merely the pool's,
and $k=3$ behaves the same way (Table~\ref{tab:ladder}). \textbf{The gain comes from ranking members
per channel, and the decaying weights are close to free either way.} A freely fitted rank profile
confirms this: it reproduces the harmonic one (correlation $0.99$, maximum deviation $0.045$) and
leaves accuracy unchanged (Appendix~\ref{app:profile}).

For practitioners without a tuning budget, a harmonic rank blend is the recommendation: \textsc{harm-5} recovers
76--87\,\% of the ensemble's gain with one ranking pass per channel and nothing fitted, equal weights on the same five cost only $0.2$\,pp, and extending the blend to the whole pool (\textsc{harm-14}) does slightly better in development and on test ($-6.30$ and $-7.89$\,\% against $-5.19$ and $-7.15$\,\%, Appendix~\ref{app:profile}).

\paragraph{2. Per-variable weights beat shared weights.} An ensemble can beat the best single model
even when one weight vector would serve the whole panel, so we fit that control: a single
simplex-constrained vector on the same pool, the same fitting data, and the same macro objective,
with each channel's statistics normalized by its own support so that dense channels do not dominate
the loss.  On test, per-variable weighting wins on every database; in development the shared vector is ahead on \textsc{UMM} and MIMIC-III.  The unshrunk per-variable fit, \textsc{simplex},
reaches $-9.16$\,\% against $-7.09$\,\% for \textsc{g-simplex}, the single shared vector; and
\textsc{ensemble}, the shrunk estimator we report throughout, reaches $-8.67$\,\%. The margin runs from $0.74$\,pp on \textsc{UMM} to $4.51$\,pp on HiRID
(Table~\ref{tab:ladder}). The advantage is smallest where the fit is most data-starved, which is
finding~6 seen from the other side. A shared vector is worth about as much as a fixed rank rule
with per-channel members ($-7.09$ against $-7.15$\,\% for \textsc{harm-5}): fitting weights globally
buys no more than ranking members locally.

\paragraph{3. The forecast combination puzzle \citep{MakridakisETAL:20, MakridakisETAL:22} does not reproduce here.} The puzzle holds that simple combinations beat estimated
ones; our fitted combiner beats every fixed rule ($-8.67$ against $-7.89$\,\% for \textsc{harm-14}, the best fixed
rule, Appendix~\ref{app:profile}). \citet{BoschETAL:25} report the same direction
at benchmark scale and explicitly describe their findings as challenging the puzzle. Two independent
settings therefore agree that learned combination wins once there is enough data to fit it, and \citet{SmithWallis:09:Puzzle} and \citet{ClaeskensETAL:16:Puzzle} trace the puzzle to the error of estimating the weights,  which makes it a question of sample size per parameter: per-channel fitting has far more effective
observations per weight than the series-level setting in which the puzzle was established. What our
data adds is the boundary condition --- \emph{how} the weights are estimated decides whether that
advantage survives. Unconstrained ridge is the clearest case: its test mean is $-4.91$\,\% against
$-8.67$\,\% for \textsc{ensemble}, and it fails outright on a different database in each protocol,
$+14.49$\,\% on MIMIC-III during development and $+1.77$\,\% on HiRID at test, both worse than a
single model. The instability is a property of unconstrained estimation on thin channels, not of one
cohort. Here the constraint and not the learning keeps the fit reliable.

\paragraph{4. Hard selection under-performs the soft combiners, and it is weak because it is
over-fitted.} Committing to the single best model on each channel beats the best single model on all
five databases, but by less than any soft combiner ($-1.48$ to $-7.30$\,\% against $-6.09$ to
$-10.95$\,\% for \textsc{ensemble}), and it is the only combiner whose advantage is ever marginal:
on eICU and \textsc{UMM} its interval nearly touches zero ($p=0.04$ on both,
Appendix~\ref{app:unc}). The weakness is a small-sample effect. Selecting on half of validation and
scoring on the disjoint half recovers only $-0.56$ to $-2.45$\,\%, against $-1.48$ to $-7.30$\,\%
when the selection sees all of validation --- HiRID moves from $-0.64$ to $-7.30$ --- though the two
are scored on different populations, so this is suggestive rather than an isolated measurement of
fitting size. An argmin over 14 noisy per-channel estimates is itself high-variance, and the modal
winner holds only 0.58--0.67 of a channel's 24 hours, with three to four distinct winners typical
(top panel of Appendix Figure~\ref{fig:maps}).

\begin{figure*}[t]
\centering
\includegraphics[width=\textwidth]{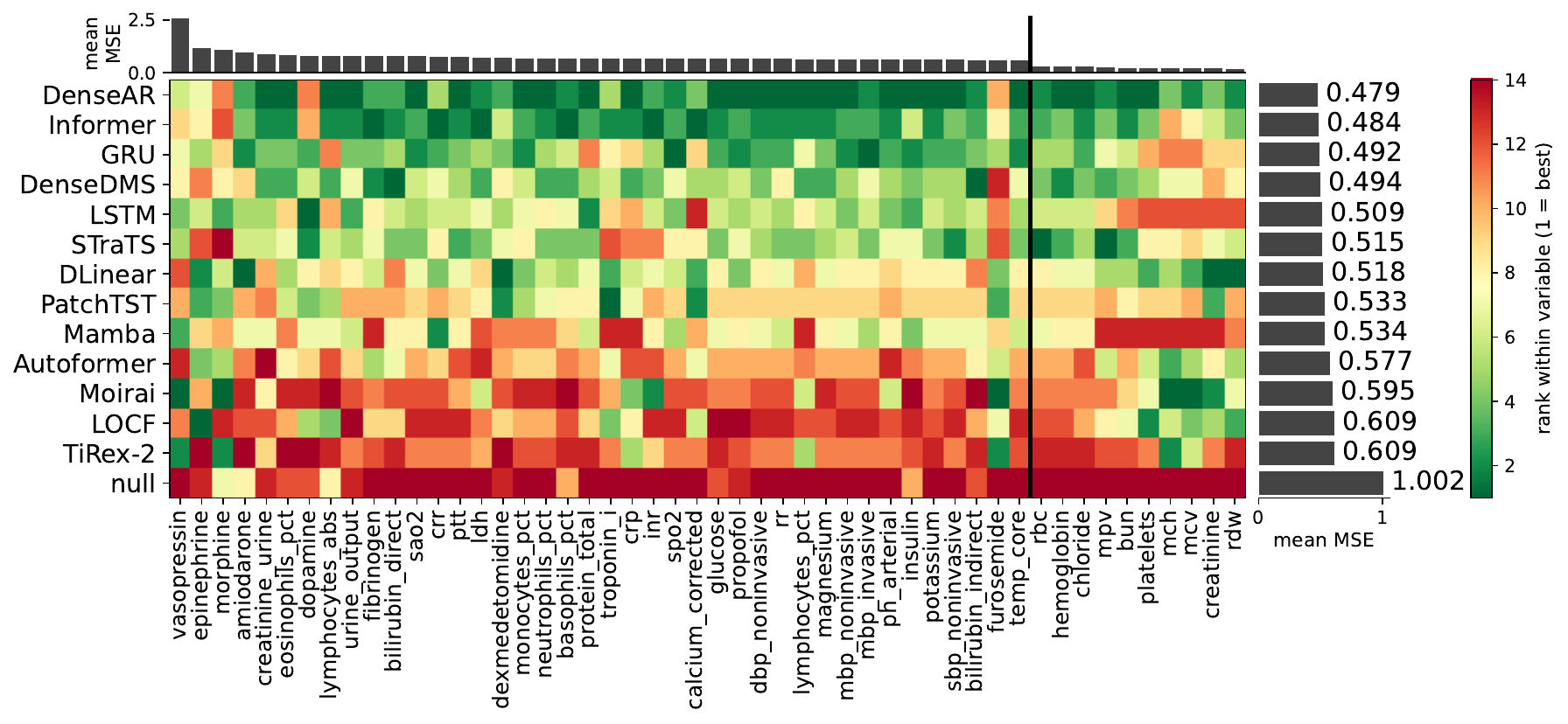}
\caption{The 14 pool members $\times$ the 40 hardest variables and, to the right of the rule,
the 10 easiest, ordered by mean MSE across the pool on validation, averaged over the
databases that chart each channel. Cells are colored by each model's \emph{rank} within that
variable; we color by rank because absolute MSE is dominated by how noisy each variable is, which
redraws the column structure and hides which model is good where. Top marginal: per-variable
difficulty. Right marginal: each model's mean MSE over all 94 variables, which sets the row order.}
\label{fig:ranked}
\end{figure*}

\paragraph{5. Performance gains in sparse channels.} Averaged over the five databases and measured
against the best member \emph{within} each system, the sparsest third of channels gains
$-11.9$\,\% against $-3.1$\,\% for the densest third; laboratory values gain $-7.5$\,\% against
$-3.7$\,\% for vital signs; and organ-failure channels gain $-7.6$\,\%. Per-system tables for
every database are in Appendix~\ref{app:system}. Sparse channels are where individual models are least reliable and therefore most
mutually complementary. All per-variable means require at least 100 observations in both the
fitting and the scoring half, which removes one channel each from \textsc{UMM} (total protein),
HiRID (vasopressin), MIMIC-III (absolute lymphocytes) and eICU (mechanical ventilation), and
none from MIMIC-IV.

\paragraph{6. Shrinkage is a small-sample correction, and the final protocol removes the small
sample.} Repeating the development protocol over five stay-grouped partitions, the gain is tight on
the three large databases (sd 0.38--0.73\,pp) and not on the two small ones: HiRID sd 2.48\,pp,
\textsc{UMM} sd 4.18\,pp, with \textsc{UMM} ranging from $-11.15$\,\% to $-1.17$\,\% on the choice
of half alone. That spread is the direct evidence that the per-channel fits are unstable at this
size, and interpolating toward the ranking prior is the correction: on validation it
wins on all five, tracking how many thin channels a database has (HiRID $+1.20$\,pp down to eICU
$+0.04$\,pp). The final fit uses the whole validation split, twice the stays, which removes the
instability and with it the reason for the correction. On test the effect is confined to one database.
Shrinkage still helps on \textsc{UMM} ($-0.26$\,pp) and costs almost nothing on MIMIC-III, MIMIC-IV
and eICU ($+0.04$, $+0.03$, $+0.21$\,pp), but on HiRID it costs $2.41$\,pp --- the whole of the
$0.49$\,pp mean difference, since excluding HiRID the shrunk and unshrunk fits are within
$0.01$\,pp. HiRID is the one database with both a poor prior (\textsc{harm-5} is $3.90$\,pp worse
than its fitted simplex, against $1.29$--$1.90$\,pp elsewhere) and enough validation stays
(1\,398) to fit without help. \textsc{greedy} selection, a quantized approximation to the same simplex
objective, tracks the unshrunk fit to within $0.04$\,pp on test (Table~\ref{tab:ladder}). We
report \textsc{ensemble} because it was fixed before test was opened.

\subsection{A stronger single model}
\label{sec:lgbm}

We extend the pool of Section~\ref{sec:pool} by a per-channel gradient-boosted tree, LightGBM \citep{KeETAL:17:LightGBM}, on hand-built window summaries, together with ARIMA \citep{BoxJenkins:70:ARIMA} and ETS \citep{GardnerETAL:85:ForecastingETS, HyndmanKhandakar:08:ETS}, and score the
extended pools once. These runs are post-hoc: the original evaluation was fixed before the
test split was used for the first time, and fixing a pool in advance does not make it
independent of the earlier test-set feedback, so we report the two evaluations separately.

The tree is fitted with a fixed 200 trees per channel, on seven summaries of that channel's own
window, one of which is the number of hours since the last real observation. The neural members
see the observation mask but never that recency as an explicit number, which is one reason the
tree does comparatively well on rarely recorded channels. It has the lowest validation error of
any member on all five databases, beats DenseAR, the strongest neural member of the original
pool, by $1.7$--$11.1\,\%$, and wins $57$--$76\,\%$ of well-supported test cells (Table~\ref{tab:winshare15}).
Table~\ref{tab:lgbm} therefore measures every combiner against LightGBM.

Per-variable stacking stays $2.9$--$5.9\,\%$ ahead of the tree, significant on four databases.
On HiRID the randomization test does not reach significance ($p = 0.11$), so we treat HiRID as inconclusive. Equal weighting of the extended pool no
longer beats the tree significantly anywhere, so the fitted per-variable weights carry the gain. Repeating
the original evaluation without Moirai and TiRex-2 changes the test gain by at most $0.15$
points, so the result does not rest on their undisclosed pretraining.

Table~\ref{tab:bymodel} gives every member and combiner. Equal weighting of the extended pool
is worse than LightGBM alone, while the stacker is $5.3\,\%$ better than it: averaging a pool
with one strong and several weaker members is not enough, the weights have to be fitted per
variable. Informer and DenseAR are effectively tied over the five databases ($+5.7$ against
$+6.1\,\%$), DenseAR leading on HiRID, MIMIC-IV and eICU and Informer on \textsc{UMM} and
MIMIC-III. The tree leads on laboratory and rarely recorded channels and is behind the best
neural member on medication by $2.4$--$9.6\,\%$, where DLinear wins on \textsc{UMM} and HiRID
and PatchTST, GRU and Informer on the larger databases.

\begin{table}[t]
\centering\small\setlength{\tabcolsep}{5pt}
\caption{Extended 15-member pool on test: change in macro-MSE relative to LightGBM, the
validation-selected best single model, in percent. Negative is better than LightGBM. Each entry is tested against LightGBM, $p$ from
approximate randomization over patients, $^{*}p<0.05$, $^{***}p<0.001$. Re-running with the
17-member pool, which adds the two foundation models, changes the stacker by at most 0.11 points.}
\label{tab:lgbm}
\begin{tabular}{lccc}
\toprule
 & stacker & harm-15 & equal \\
\midrule
\textsc{UMM}  & $-2.85^{*}$   & $-1.42$       & $+2.32$ \\
HiRID         & $-5.27$       & $-4.38$       & $+0.77$ \\
MIMIC-III     & $-4.97^{***}$ & $-3.82^{***}$ & $+0.22$ \\
MIMIC-IV      & $-5.90^{***}$ & $-4.36^{***}$ & $+0.04$ \\
eICU          & $-4.89^{***}$ & $-4.02^{***}$ & $-0.81$ \\
\bottomrule
\end{tabular}

\end{table}

\subsection{Which model wins where}
\label{sec:which}

Counting well-supported (variable, hour) test cells won by each family
(Table~\ref{tab:winshare}) shows that the leader differs by database. Figure~\ref{fig:ranked} shows each member's rank on every variable.

\begin{table}[t]
\centering\small\setlength{\tabcolsep}{4pt}
\caption{Share of well-supported (variable, hour) test cells won by each family (\%). Column
subheadings give training stays. Shares sum to 100\,\% per column; every member is listed in
Appendix~\ref{app:win}.}
\label{tab:winshare}
\begin{tabular}{lrrrrr}
\toprule
family & \textsc{UMM} & HiRID & M-III & M-IV & eICU \\
 & \scriptsize 2.9k & \scriptsize 5.5k & \scriptsize 18k & \scriptsize 30k & \scriptsize 50k \\
\midrule
STraTS & 12.2 & \textbf{24.4} & 30.0 & \textbf{40.0} & 35.5 \\
DenseAR & \textbf{22.8} & 21.1 & \textbf{34.2} & 20.6 & \textbf{35.8} \\
Informer & 21.2 & 9.3 & 7.1 & 7.5 & 2.9 \\
GRU & 6.2 & 11.1 & 2.6 & 14.1 & 3.9 \\
Dense DMS & 9.1 & 8.7 & 7.9 & 3.0 & 7.3 \\
DLinear & 13.9 & 16.5 & 1.7 & 1.1 & 0.8 \\
PatchTST & 5.7 & 2.9 & 7.1 & 5.2 & 5.6 \\
LOCF + null & 3.4 & 2.5 & 6.2 & 4.5 & 3.1 \\
Moirai + TiRex-2 & 3.2 & 2.4 & 2.5 & 2.3 & 1.5 \\
LSTM & 1.7 & 0.5 & 0.4 & 0.9 & 2.7 \\
Autoformer & 0.3 & 0.6 & 0.3 & 0.1 & 0.7 \\
Mamba & 0.3 & 0.1 & 0.0 & 0.7 & 0.2 \\
\bottomrule
\end{tabular}
\end{table}

STraTS wins $12\,\%$ of cells on \textsc{UMM} and $40\,\%$ on MIMIC-IV; DLinear moves the
opposite way, $13.9\,\%$ against $0.8\,\%$; DenseAR is the most consistent member at
$20.6$--$35.8\,\%$. These patterns are associated with database size, but size co-varies with
case mix, measurement practice and tuning outcome, so they do not show that an architecture
needs more data. On channels that behave like random walks over 24 hours, \textsc{locf} and the
training mean are the right model and win $2.5$--$6.2\,\%$ of cells. Members differ mainly by
training regime rather than by architecture family: hyperparameter variants of one decoder
correlate at $\rho \geq 0.965$, transformer variants span the whole cross-family range, and the
two zero-shot models are the least correlated learned members (Appendix~\ref{app:div}).

\section{Discussion}
\label{sec:discussion}

Most of the improvement comes from choosing members per variable rather than from fitting their
weights precisely. A fixed rank-based blend of all fourteen members recovers about $90\,\%$ of
the stacker's gain, and fitted weights add a further $0.3$--$1.3$ points. Fitting matters more
as data grow. Random convex weights still beat the best single model on  four databases, by $1.1$--$6.8\,\%$ on average, but not on MIMIC-IV ($+0.5\,\%$), where the fitted stacker gains $6.5\,\%$ (Table~\ref{tab:controls}).

The gain needs different models. Stacking four seeds of a single architecture gains
$1$--$8\,\%$ over its best seed, while the stacked diverse pool has a $4$--$14\,\%$ lower error than the stacked DenseAR seeds (Table~\ref{tab:controls}). Random subsets
of the pool recover about nine-tenths of the gain with four to six members
(Figure~\ref{fig:poolsize}); the only subsets that lose to the best single model are individual
two-member draws, so pool spread matters more than pool size.

Adding a gradient-boosted tree raises the single-model baseline substantially and the
combination still improves on it. Which model is best stays a per-variable question, and it is
not a contest between the tree and the neural members: DLinear wins about $8\,\%$ of cells on
the two small databases and the medication channels there, \textsc{locf} wins $1$--$3\,\%$, and
the all-zeros baseline wins $2.8\,\%$ on MIMIC-III in the extended pool (Table~\ref{tab:winshare15}).

The target is the value a clinician recorded, under the measurement practice of each hospital.
Sampling is informative, since a laboratory test is often ordered because deterioration is
suspected, so gains on sparse channels are gains on a benchmark and not on unobserved
physiology. How forecast error propagates into SOFA and SAPS-II is beyond the scope of this
paper. Each database is evaluated within itself, so the five results are replications of one
procedure. Validation is used twice, to select base checkpoints and to fit the combiner, which
makes development numbers mildly optimistic; the held-out result is unaffected because the
combiner was fixed before test was used.

\section{Conclusion}
Forecast combination is the more durable finding of this study. Which architecture leads changed
completely when we added a gradient-boosted tree, and it changes from variable to variable
within every database, but combining several models per variable helped on all five databases,
against both the original pool's best member and the much stronger tree. A ranking of
architectures is therefore the less transferable result to report, and per-variable combination
is cheap enough to be a default rather than a contribution: member predictions are cached once,
after which the entire combiner ladder over five databases is a seven-minute CPU job, and
random subsets of four to six members recover about nine-tenths of the gain (Figure~\ref{fig:poolsize}). We would report
per-variable combination as a baseline in clinical forecasting work, in the same way a no-change
predictor is reported.

\acks{The authors acknowledge support by the state of Baden-W\"urttemberg through bwHPC and the
German Research Foundation (DFG) through grant INST 35/1597-1 FUGG. 

Marius Fracarolli is funded by the Helmholtz Information \& Data Science School for
Health (HIDSS4Health), Karlsruhe/Heidelberg, Germany. 

We thank the clinical staff of the
intensive care unit at University Medicine Mannheim for the records analyzed here, and the
PhysioNet maintainers for curating the four public databases.}

\bibliography{references}

@article{GanaieETAL:21,
title = {Ensemble deep learning: A review},
journal = {Engineering Applications of Artificial Intelligence},
volume = {115},
pages = {105151},
year = {2022},
issn = {0952-1976},
url = {https://www.sciencedirect.com/science/article/pii/S095219762200269X},
author = {M.A. Ganaie and Minghui Hu and A.K. Malik and M. Tanveer and P.N. Suganthan}
}

@article{HansenSalamon:90,
  author =	 {Hansen, L.K. and Salamon, P.},
  journal =	 {{IEEE} Transactions on Pattern Analysis and Machine
                  Intelligence},
  title =	 {Neural network ensembles},
  year =	 1990,
  volume =	 12,
  number =	 10,
  pages =	 {993-1001},
  url = {https://dl.acm.org/doi/10.1109/34.58871}
}

@Article{HewamalageETAL:23,
  author = 	 {Hansika Hewamalage and Klaus Ackermann and Christoph Bergmeir},
  title = 	 {Forecast evaluation for data scientists: common pitfalls and best practices},
  journal = 	 {Data Mining and Knowledge Discovery},
  year = 	 2023,
  volume = 	 37,
  pages = 	 {788-832},
  url = {https://doi.org/10.1007/s10618-022-00894-5}
}

@article{HochreiterSchmidhuber:97,
  author =	 {Hochreiter, Sepp and Schmidhuber, J\"{u}rgen},
  title =	 {Long Short-Term Memory ({LSTM})},
  journal =	 {Neural Computation},
  volume =	 9,
  number =	 8,
  year =	 1997,
  pages =	 {1735--1780},
  url =		 {http://dx.doi.org/10.1162/neco.1997.9.8.1735},
  doi =		 {10.1162/neco.1997.9.8.1735},
}

@Article{HylandETAL:20,
  author =	 {Hyland, Stephanie L.  and Faltys, Martin and
                  H\"{u}ser, Matthias and Lyu, Xinrui and Gumbsch,
                  Thomas and Esteban, Crist\'{o}bal and Bock,
                  Christian and Horn, Max and Moor, Michael and Rieck,
                  Bastian and Zimmermann, Marc and Bodenham, Dean and
                  Borgwardt, Karsten and R\"{a}tsch, Gunnar and Merz,
                  Tobias M.},
  title =	 {Early prediction of circulatory failure in the
                  intensive care unit using machine learning},
  journal =	 {Nature Medicine},
  year =	 2020,
  volume =	 26,
  pages =	 {364-373},
  url =		 {https://doi.org/10.1038/s41591-020-0789-4}
}

@Article{JohnsonETAL:16,
  author =	 {Alistair E.W. Johnson and Tom J. Pollard and Lu Shen
                  and Li-wei H. Lehman and Mengling Feng and Mohammad
                  Ghassemi and Benjamin Moody and Peter Szolovits and
                  Leo Anthony Celi and Roger G. Mark},
  year =	 2016,
  pages =	 160035,
  title =	 {{MIMIC-III}, a freely accessible critical care
                  database},
  volume =	 3,
  number =	 1,
  journal =	 {Scientific Data},
  url =		 {https://doi.org/10.1038/sdata.2016.35},
}

@Article{JohnsonETAL:23,
  author =	 {Johnson, Alistair E. W.  and Bulgarelli, Lucas and
                  Shen, Lu and Gayles, Alvin and Shammout, Ayad and
                  Horng, Steven and Pollard, Tom J.  and Hao, Sicheng
                  and Moody, Benjamin and Gow, Brian and Lehman,
                  Li-wei H.  and Celi, Leo A.  and Mark, Roger G.},
  title =	 {{MIMIC-IV}, a freely accessible electronic health
                  record dataset},
  journal =	 {Scientific Data},
  year =	 2023,
  volume =	 10,
  number =	 1,
  url =		 {https://doi.org/10.1038/s41597-022-01899-x}
}

@inproceedings{NieETAL:23,
  title =	 {A Time Series is Worth 64 Words: Long-term
                  Forecasting with Transformers},
  author =	 {Yuqi Nie and Nam H Nguyen and Phanwadee Sinthong and
                  Jayant Kalagnanam},
  booktitle =	 {Proceedings of the 11th International Conference on
                  Learning Representations {(ICLR)}},
  year =	 2023,
  url =		 {https://openreview.net/forum?id=Jbdc0vTOcol},
  address =	 {Kigali, Rwanda}
}

@InCollection{PerroneCooper:92,
  author =	 {Michael P. Perrone and Leon N. Cooper},
  title =	 {When Networks Disagree: Ensemble Methods for Hybrid
                  Neural Networks},
  booktitle =	 {Artificial Neural Networks for Speech and Image
                  Processing},
  publisher =	 {Chapman-Hall},
  year =	 1992,
  editor =	 {R. J. Mammone},
  pages =	 {126-142},
  address =	 {New York},
  url = {https://www.worldscientific.com/doi/10.1142/9789812795885_0025}
}

@Article{PollardETAL:18,
  author =	 {Tom J. Pollard and Alistair E. W. Johnson and Jesse
                  D. Raffa and Leo A. Celi and Roger G. Mark and Omar
                  Badawi},
  title =	 {The {eICU} Collaborative Research Database, a freely
                  available multi-center database for critical care
                  research},
  journal =	 {Scientific Data},
  year =	 2018,
  volume =	 5,
  number =	 180178,
  url =		 {https://doi.org/10.1038/sdata.2018.178}
}

@Article{SchamoniETAL:22,
  author =	 {Shigehiko Schamoni and Michael Hagmann and Stefan
                  Riezler},
  title =	 {Ensembling Neural Networks for Improved Prediction
                  and Privacy in Early Diagnosis of Sepsis},
  journal =	 {Proceedings of Machine Learning Research: Machine
                  Learning for Healthcare},
  year =	 2022,
  volume =	 182,
  pages =	 {1-23},
  url =
                  {https://proceedings.mlr.press/v182/schamoni22a/schamoni22a.pdf}
}

@Article{StaniekETAL:24mlhc,
  author =	 {Michael Staniek and Marius Fracarolli and Michael
                  Hagmann and Stefan Riezler},
  title =	 {Early Prediction of Causes (not Effects) in
                  Healthcare by Long-Term Clinical Time Series
                  Forecasting},
  journal =	 {Proceedings of Machine Learning Research: Machine
                  Learning for Healthcare},
  year =	 2024,
  volume =	 252,
  pages =	 {1-29},
  url={https://proceedings.mlr.press/v252/staniek24a.html}
}

@article{TipirneniReddy:21,
  author =	 {Sindhu Tipirneni and Chandan K. Reddy},
  title =	 {Self-Supervised Transformer for Sparse and
                  Irregularly Sampled Multivariate Clinical
                  Time-Series},
  year =	 2022,
  volume =	 16,
  number =	 6,
  url =		 {https://doi.org/10.1145/3516367},
  journal =	 {{ACM} Transactions on Knowledge Discovery from Data},
}

@article{Wolpert:92,
title = {Stacked generalization},
journal = {Neural Networks},
volume = {5},
number = {2},
pages = {241-259},
year = {1992},
issn = {0893-6080},
url = {https://www.sciencedirect.com/science/article/pii/S0893608005800231},
author = {David H. Wolpert}
}

@inproceedings{WuETAL:21,
  author =	 {Wu, Haixu and Xu, Jiehui and Wang, Jianmin and Long,
                  Mingsheng},
  booktitle =	 {Advances in Neural Information Processing Systems
                  {(NeurIPS)}},
  title =	 {Autoformer: Decomposition Transformers with
                  Auto-Correlation for Long-Term Series Forecasting},
  url =
                  {https://proceedings.neurips.cc/paper_files/paper/2021/file/bcc0d400288793e8bdcd7c19a8ac0c2b-Paper.pdf},
  year =	 2021,
  address =	 {virtual}
}

@inproceedings{ZengETAL:23,
  title =	 {Are transformers effective for time series
                  forecasting?},
  author =	 {Zeng, Ailing and Chen, Muxi and Zhang, Lei and Xu,
                  Qiang},
  booktitle =	 {Proceedings of the 37th {AAAI} Conference on
                  Artificial Intelligence},
  year =	 2023,
  address =	 {Washington {DC, USA}},
  url =
                  {https://ojs.aaai.org/index.php/AAAI/article/view/26317/26089}
}

@inproceedings{ZhouETAL:21,
  author =	 {Haoyi Zhou and Shanghang Zhang and Jieqi Peng and
                  Shuai Zhang and Jianxin Li and Hui Xiong and Wancai
                  Zhang},
  title =	 {Informer: Beyond Efficient Transformer for Long
                  Sequence Time-Series Forecasting},
  booktitle =	 {The Thirty-Fifth {AAAI} Conference on Artificial
                  Intelligence},
  address =	 {virtual},
  year =	 2021,
  url =		 { https://doi.org/10.1609/aaai.v35i12.17325 },
}

@article{HaugsdalETAL:23:Persistence,
  title   = {Persistence Initialization: a novel adaptation of the Transformer architecture for time series forecasting},
  author  = {Haugsdal, Espen and Aune, Erlend and Ruocco, Massimiliano},
  journal = {Applied Intelligence},
  volume  = {53},
  number  = {22},
  pages   = {26781--26796},
  year    = {2023},
  url     = {https://doi.org/10.1007/s10489-023-04927-4}
}

@article{MakridakisETAL:20,
title = {The {M4} Competition: 100,000 time series and 61 forecasting methods},
journal = {International Journal of Forecasting},
volume = {36},
number = {1},
pages = {54-74},
year = {2020},
note = {M4 Competition},
issn = {0169-2070},
url = {https://www.sciencedirect.com/science/article/pii/S0169207019301128},
author = {Spyros Makridakis and Evangelos Spiliotis and Vassilios Assimakopoulos}
}

@article{MakridakisETAL:22,
title = {M5 accuracy competition: Results, findings, and conclusions},
journal = {International Journal of Forecasting},
volume = {38},
number = {4},
pages = {1346-1364},
year = {2022},
note = {Special Issue: M5 competition},
issn = {0169-2070},
url = {https://www.sciencedirect.com/science/article/pii/S0169207021001874},
author = {Spyros Makridakis and Evangelos Spiliotis and Vassilios Assimakopoulos}
}

@article{AnsariETAL:24:Chronos,
  title={Chronos: Learning the language of time series},
  author={Ansari, Abdul Fatir and Stella, Lorenzo and Turkmen, Caner and Zhang, Xiyuan and Mercado, Pedro and Shen, Huibin and Shchur, Oleksandr and Rangapuram, Syama Sundar and Arango, Sebastian Pineda and Kapoor, Shubham and others},
  journal={Transactions on Machine Learning Research},
  year={2024},
  url={https://arxiv.org/abs/2403.07815},
}

@inproceedings{KroghVedelsby:95,
 author = {Krogh, Anders and Vedelsby, Jesper},
 booktitle = {Advances in Neural Information Processing Systems},
 editor = {G. Tesauro and D. Touretzky and T. Leen},
 pages = {},
 publisher = {MIT Press},
 title = {Neural Network Ensembles, Cross Validation, and Active Learning},
 url = {https://proceedings.neurips.cc/paper_files/paper/1994/file/b8c37e33defde51cf91e1e03e51657da-Paper.pdf},
 volume = {7},
 year = {1994}
}

@article{Breiman:96stacked,
  title   = {Stacked Regressions},
  author  = {Breiman, Leo},
  journal = {Machine Learning},
  volume  = {24},
  number  = {1},
  pages   = {49--64},
  year    = {1996},
  url = {https://link.springer.com/article/10.1007/BF00117832}
}

@article{BatesGranger:69,
  title   = {The Combination of Forecasts},
  author  = {Bates, J. M. and Granger, C. W. J.},
  journal = {Journal of the Operational Research Society},
  volume  = {20},
  number  = {4},
  pages   = {451--468},
  year    = {1969},
  url     = {https://link.springer.com/article/10.1057/jors.1969.103}
}

@inproceedings{CaruanaETAL:04,
  title     = {Ensemble Selection from Libraries of Models},
  author    = {Caruana, Rich and Niculescu-Mizil, Alexandru and Crew, Geoff and Ksikes, Alex},
  booktitle = {Proceedings of the Twenty-first International Conference on Machine Learning (ICML)},
  year      = {2004},
  url = {https://dl.acm.org/doi/10.1145/1015330.1015432}
}

@inproceedings{DuchiETAL:08,
  title     = {Efficient Projections onto the $\ell_1$-Ball for Learning in High Dimensions},
  author    = {Duchi, John and Shalev-Shwartz, Shai and Singer, Yoram and Chandra, Tushar},
  booktitle = {Proceedings of the 25th International Conference on Machine Learning (ICML)},
  year      = {2008},
  url = {https://dl.acm.org/doi/10.1145/1390156.1390191}
}

@inproceedings{GuDao:23,
  title         = {Mamba: Linear-Time Sequence Modeling with Selective State Spaces},
  author        = {Gu, Albert and Dao, Tri},
  journal       = {arXiv preprint arXiv:2312.00752},
  eprint        = {2312.00752},
  archivePrefix = {arXiv},
  url           = {https://arxiv.org/abs/2312.00752},
  booktitle = {First Conference on Language Modeling (COLM)},
  year = {2024}
}

@inproceedings{WooETAL:24:Moirai,
author = {Woo, Gerald and Liu, Chenghao and Kumar, Akshat and Xiong, Caiming and Savarese, Silvio and Sahoo, Doyen},
title = {Unified training of universal time series forecasting transformers},
year = {2024},
publisher = {JMLR.org},
booktitle = {Proceedings of the 41st International Conference on Machine Learning},
articleno = {2178},
numpages = {25},
location = {Vienna, Austria},
series = {ICML'24},
url = {https://dl.acm.org/doi/10.5555/3692070.3694248}
}

@misc{PodestETAL:26:TiRex2,
  author =       {Patrick Podest and Marco Pichler and Elias B\"urger and
                  Levente Z\'olyomi and Bernhard Voggenberger and Wilhelm
                  Berghammer and Daniel Klotz and Sebastian B\"ock and
                  G\"unter Klambauer and Sepp Hochreiter},
  title =        {{TiRex-2}: Generalizing {TiRex} to Multivariate Data and
                  Streaming},
  year =         2026,
  eprint =       {2607.01204},
  archivePrefix = {arXiv},
  primaryClass = {cs.LG},
  url =          {https://arxiv.org/abs/2607.01204}
}

@InProceedings{BoschETAL:25,
  title = 	 {Multi-layer Stack Ensembles for Time Series Forecasting},
  author =       {Bosch, Nathanael and Shchur, Oleksandr and Erickson, Nick and Bohlke-Schneider, Michael and Turkmen, Ali Caner},
  booktitle = 	 {Proceedings of the Fourth International Conference on Automated Machine Learning},
  pages = 	 {15/1--30},
  year = 	 {2025},
  editor = 	 {Akoglu, Leman and Doerr, Carola and van Rijn, Jan N. and Garnett, Roman and Gardner, Jacob R.},
  volume = 	 {293},
  series = 	 {Proceedings of Machine Learning Research},
  month = 	 {08--11 Sep},
  publisher =    {PMLR},
  url = 	 {https://proceedings.mlr.press/v293/bosch25a.html}
}

@inproceedings{CormackETAL:09:ReciprocalRank,
author = {Cormack, Gordon V. and Clarke, Charles L A and Buettcher, Stefan},
title = {Reciprocal rank fusion outperforms condorcet and individual rank learning methods},
year = {2009},
isbn = {9781605584836},
publisher = {Association for Computing Machinery},
address = {New York, NY, USA},
url = {https://doi.org/10.1145/1571941.1572114},
booktitle = {Proceedings of the 32nd International ACM SIGIR Conference on Research and Development in Information Retrieval},
pages = {758–759},
numpages = {2},
location = {Boston, MA, USA},
series = {SIGIR '09}
}

@article{HoerlKennard:70:RidgeRegression,
 ISSN = {00401706},
 URL = {http://www.jstor.org/stable/1271436},
 author = {Arthur E. Hoerl and Robert W. Kennard},
 journal = {Technometrics},
 number = {1},
 pages = {55--67},
 publisher = {[Taylor & Francis, Ltd., American Statistical Association, American Society for Quality]},
 title = {Ridge Regression: Biased Estimation for Nonorthogonal Problems},
 urldate = {2026-09-09},
 volume = {12},
 year = {1970}
}

@inproceedings{ChoETAL:14:GRU,
    title = "Learning Phrase Representations using {RNN} Encoder{--}Decoder for Statistical Machine Translation",
    author = {Cho, Kyunghyun  and
      van Merri{\"e}nboer, Bart  and
      Gulcehre, Caglar  and
      Bahdanau, Dzmitry  and
      Bougares, Fethi  and
      Schwenk, Holger  and
      Bengio, Yoshua},
    editor = "Moschitti, Alessandro  and
      Pang, Bo  and
      Daelemans, Walter",
    booktitle = "Proceedings of the 2014 Conference on Empirical Methods in Natural Language Processing ({EMNLP})",
    month = oct,
    year = "2014",
    address = "Doha, Qatar",
    publisher = "Association for Computational Linguistics",
    url = "https://aclanthology.org/D14-1179/",
    doi = "10.3115/v1/D14-1179",
    pages = "1724--1734"
}

@article{SmithWallis:09:Puzzle,
author = {Smith, Jeremy and Wallis, Kenneth F.},
title = {A Simple Explanation of the Forecast Combination Puzzle},
journal = {Oxford Bulletin of Economics and Statistics},
volume = {71},
number = {3},
pages = {331-355},
doi = {https://doi.org/10.1111/j.1468-0084.2008.00541.x},
url = {https://onlinelibrary.wiley.com/doi/abs/10.1111/j.1468-0084.2008.00541.x},
eprint = {https://onlinelibrary.wiley.com/doi/pdf/10.1111/j.1468-0084.2008.00541.x},
year = {2009}
}

@article{ClaeskensETAL:16:Puzzle,
title = {The forecast combination puzzle: A simple theoretical explanation},
journal = {International Journal of Forecasting},
volume = {32},
number = {3},
pages = {754-762},
year = {2016},
issn = {0169-2070},
doi = {https://doi.org/10.1016/j.ijforecast.2015.12.005},
url = {https://www.sciencedirect.com/science/article/pii/S0169207016000327},
author = {Gerda Claeskens and Jan R. Magnus and Andrey L. Vasnev and Wendun Wang}
}

@inproceedings{KeETAL:17:LightGBM,
 author = {Ke, Guolin and Meng, Qi and Finley, Thomas and Wang, Taifeng and Chen, Wei and Ma, Weidong and Ye, Qiwei and Liu, Tie-Yan},
 booktitle = {Advances in Neural Information Processing Systems},
 editor = {I. Guyon and U. Von Luxburg and S. Bengio and H. Wallach and R. Fergus and S. Vishwanathan and R. Garnett},
 pages = {},
 publisher = {Curran Associates, Inc.},
 title = {LightGBM: A Highly Efficient Gradient Boosting Decision Tree},
 url = {https://proceedings.neurips.cc/paper_files/paper/2017/file/6449f44a102fde848669bdd9eb6b76fa-Paper.pdf},
 volume = {30},
 year = {2017}
}

@article{HyndmanKhandakar:08:ETS,
title={Automatic time series forecasting: the forecast package for R},
author={Hyndman, Rob J and Khandakar, Yeasmin},
journal={Journal of statistical software},
volume={27},
pages={1--22},
year={2008}
}

@book{BoxJenkins:70:ARIMA,
  title     = {Time Series Analysis: Forecasting and Control},
  author    = {Box, George E. P. and Jenkins, Gwilym M.},
  year      = {1970},
  publisher = {Holden-Day},
  address   = {San Francisco}
}

@article{GardnerETAL:85:ForecastingETS,
  author    = {Gardner, Jr., Everette S. and McKenzie, Ed.},
  title     = {Forecasting Trends in Time Series},
  journal   = {Management Science},
  year      = {1985},
  volume    = {31},
  number    = {10},
  pages     = {1237--1246},
  month     = oct,
  publisher = {INFORMS},
  doi       = {10.1287/mnsc.31.10.1237},
  url       = {https://www.bauer.uh.edu/egardner/3301H%20Operations%20Management/ESG%20Publications/1985%20Forecasting%20trends.pdf}
}

\appendix

\counterwithin{table}{section}
\counterwithin{figure}{section}
\renewcommand{\thetable}{\thesection\arabic{table}}
\renewcommand{\theHtable}{\thesection\arabic{table}}
\renewcommand{\thefigure}{\thesection\arabic{figure}}
\renewcommand{\theHfigure}{\thesection\arabic{figure}}

\onecolumn

\section{Databases}\label{app:data}

Sizes are in Table~\ref{tab:data}. \emph{Charted} counts the channels a database records at
least once inside a 24-hour target window, and is the denominator $|C|$ of the metric;
never-recorded channels are excluded from every per-variable mean. It differs by at most one from
the non-dash entries of Table~\ref{tab:coverage}, which counts a channel recorded anywhere in a
stay --- a channel seen only outside the first seven days, or only in the input half of a window,
appears in one and not the other. \emph{Density} is the fraction
of (variable, hour) target cells carrying an observation, over the full training split.

Windows are drawn from the first seven ICU days at a 4-hour stride, each with a 24-hour input and a
24-hour horizon, giving up to 31 windows per stay. Splits are 64 train /16 validation/20 test by patient at one fixed seed;
Table~\ref{tab:splits} gives the resulting sizes.

\begin{table}[htbp]\centering\small
\caption{Database sizes by split, and the mean number of observed (variable, hour) cells per
24-hour target window out of $24\times94=2256$ possible.}\label{tab:splits}
\begin{tabular}{lrrrrrrr}
\toprule
 & \multicolumn{2}{c}{train} & \multicolumn{2}{c}{val} & \multicolumn{2}{c}{test} & obs./ \\
 & stays & windows & stays & windows & stays & windows & window \\
\midrule
MIMIC-III & 18\,242 & 274\,195 & 4\,568 & 69\,351 & 5\,704 & 87\,565 & 249 \\
MIMIC-IV & 29\,604 & 444\,870 & 7\,382 & 111\,567 & 9\,234 & 137\,111 & 257 \\
eICU & 49\,533 & 685\,134 & 12\,197 & 168\,643 & 15\,385 & 211\,251 & 200 \\
HiRID & 5\,543 & 88\,407 & 1\,398 & 22\,143 & 1\,680 & 26\,133 & 304 \\
\textsc{UMM} & 2\,889 & 62\,466 & 710 & 15\,054 & 891 & 19\,541 & 425 \\
\bottomrule
\end{tabular}
\end{table}

\noindent The last column is the mean number of observed (variable, hour) cells per 24-hour target
window, out of $24\times94=2256$ possible. Even \textsc{UMM}, the densest, observes under a fifth.

\paragraph{Variable systems.} Results below are reported over the catalog grouping (\emph{lab},
\emph{vital}, \emph{medication}, \emph{ventilation}), over the organ-failure channels
(\emph{sofa\_relevant}), and over observation-frequency tertiles computed within each database
(\emph{freq:low/mid/high}). Group means use only channels with at least 100 observations: one HiRID
medication channel with twelve observed cells otherwise supplied 41\% of that group's mean.

\paragraph{Per-channel coverage.} Table~\ref{tab:coverage} gives the percentage of stays in
which each channel is recorded at least once. Coverage varies with local practice, not data volume:
albumin appears in 83\% of \textsc{UMM} stays and 14.8\% of HiRID stays. Hence the
$\geq 100$-observation filter on the group-level tables.

\begin{table}[htbp]
\centering\footnotesize
\caption{Percentage of ICU stays in which each channel is recorded at least once. A dash marks a channel the database never records. Channels are ordered by mean coverage. Coverage spans the full range within every database, which is why all per-variable means here carry a $\geq 100$-observation filter; the same data is plotted in Figure~\ref{fig:coverage}.}\label{tab:coverage}

\begin{tabular}{lrrrrr|lrrrrr}
\toprule
channel & \textsc{UMM} & HiRID & M-III & M-IV & eICU & channel  & \textsc{UMM}  & HiRID  & M-III  & M-IV  & eICU  \\
\midrule
hr & 99.9 & 100.0 & 99.4 & 99.9 & 98.7 & peep & 74.2 & 66.3 & 44.7 & 39.8 & 31.3 \\
spo2 & 99.8 & 100.0 & 99.3 & 99.9 & 97.7 & alp & 82.6 & 36.9 & 36.8 & 46.2 & 45.1 \\
glucose & 99.7 & 99.3 & 99.0 & 98.6 & 96.5 & tidal\_volume & 70.9 & 66.1 & 40.7 & 40.1 & 26.2 \\
potassium & 99.8 & 98.5 & 99.1 & 98.6 & 95.3 & albumin & 83.0 & 14.8 & 32.6 & 31.7 & 50.7 \\
sodium & 99.8 & 98.4 & 99.0 & 98.6 & 94.9 & neutrophils\_pct & 38.1 & 59.1 & 27.0 & 25.1 & 54.1 \\
hemoglobin & 99.8 & 97.5 & 98.6 & 98.4 & 93.4 & pt & --- & --- & 78.7 & 81.6 & 34.5 \\
wbc & 98.3 & 97.5 & 98.8 & 98.4 & 92.9 & amv & 71.9 & --- & 44.4 & 40.0 & 15.6 \\
creatinine & 98.0 & 94.9 & 99.0 & 98.6 & 94.7 & svo2 & 77.3 & 33.7 & 30.8 & 18.6 & --- \\
platelets & 98.2 & 97.1 & 98.8 & 98.4 & 92.3 & lymphocytes\_pct & 38.1 & --- & 27.1 & 25.1 & 60.0 \\
mcv & 98.3 & 97.5 & 98.6 & 98.4 & 90.6 & monocytes\_pct & 38.1 & --- & 27.0 & 25.1 & 59.6 \\
mchc & 98.3 & 97.2 & 98.6 & 98.4 & 90.6 & eosinophils\_pct & 38.1 & --- & 27.0 & 25.1 & 57.5 \\
hco3 & 97.1 & 91.1 & 98.9 & 98.6 & 93.8 & ldh & 82.9 & --- & 28.4 & 33.9 & --- \\
chloride & 97.1 & 89.5 & 98.9 & 98.6 & 94.8 & basophils\_pct & 38.1 & --- & 27.0 & 25.1 & 54.4 \\
mch & 98.3 & 97.5 & 98.6 & 98.4 & 86.3 & weight & --- & 5.9 & 59.1 & 75.7 & --- \\
bun & 98.1 & 85.1 & 99.0 & 98.6 & 94.7 & mechanical\_vent. & 76.1 & --- & 17.8 & 36.6 & --- \\
rr & 71.2 & 100.0 & 99.4 & 99.8 & 92.9 & fibrinogen & 69.5 & 13.7 & 13.7 & 15.8 & 5.0 \\
temp\_core & 99.3 & 65.2 & 99.2 & 99.3 & 95.9 & propofol & 18.5 & 57.4 & 10.1 & 25.4 & 4.7 \\
urine\_output & 99.5 & 98.6 & 93.9 & 96.6 & 67.9 & insulin & 50.3 & 47.2 & 4.5 & 8.5 & 4.4 \\
gcs\_eye & 65.2 & 100.0 & 99.4 & 99.9 & 64.9 & norepinephrine & 42.4 & 37.9 & 6.1 & 20.0 & 6.1 \\
gcs\_motor & 64.8 & 100.0 & 99.4 & 99.8 & 64.9 & lipase & 70.7 & 9.3 & 17.1 & 9.8 & --- \\
gcs\_verbal & 64.6 & 100.0 & 99.4 & 99.9 & 64.9 & crp & 97.9 & --- & 1.0 & 3.1 & --- \\
magnesium & 73.0 & 85.0 & 98.0 & 98.2 & 67.7 & etco2 & 9.3 & 67.4 & --- & 15.1 & 4.6 \\
phosphate & 71.3 & 84.6 & 94.3 & 97.6 & 49.4 & crr & --- & --- & 2.2 & 88.7 & --- \\
hematocrit & 98.3 & --- & 98.9 & 98.4 & 93.5 & midazolam & 34.4 & 39.1 & 6.7 & 8.4 & 1.9 \\
inr & 97.8 & 66.8 & 78.7 & 81.6 & 35.4 & cardiac\_output & 55.4 & 25.0 & --- & 2.1 & 1.9 \\
ph\_arterial & 97.1 & 82.0 & 63.9 & 57.0 & 34.7 & lymphocytes\_abs & --- & 59.0 & 0.3 & 13.7 & --- \\
sbp\_noninvasive & 23.8 & 45.7 & 91.4 & 95.1 & 76.3 & mpv & --- & --- & --- & --- & 63.9 \\
dbp\_noninvasive & 23.8 & 45.7 & 91.4 & 95.1 & 76.3 & pct & 58.2 & 4.1 & --- & --- & --- \\
mbp\_noninvasive & 23.7 & 45.7 & 91.3 & 95.2 & 76.3 & stroke\_volume & 55.5 & --- & --- & 1.6 & --- \\
base\_excess & 97.1 & 91.1 & 58.1 & 50.9 & 29.8 & protein\_total & 1.6 & --- & 2.2 & 2.1 & 45.7 \\
mbp\_invasive & 96.5 & 98.7 & 59.1 & 43.1 & 27.0 & fentanyl & 1.2 & 41.3 & 2.7 & 1.7 & 3.8 \\
dbp\_invasive & 96.5 & 98.7 & 59.1 & 43.0 & 26.9 & heparin & 22.8 & --- & 5.4 & 14.5 & 2.3 \\
sbp\_invasive & 96.5 & 98.7 & 59.1 & 43.0 & 26.9 & troponin\_i & 31.9 & --- & 1.6 & --- & 9.3 \\
pao2 & 97.4 & 81.9 & 58.1 & 50.9 & 35.3 & troponin\_t & --- & 30.6 & --- & 11.6 & --- \\
paco2 & 97.4 & 81.8 & 58.1 & 50.9 & 35.0 & bilirubin\_direct & 5.0 & 9.8 & 6.1 & 6.3 & 11.5 \\
calcium\_total & 18.2 & 14.1 & 93.6 & 97.5 & 92.4 & creatinine\_urine & --- & --- & 14.3 & 11.8 & 3.0 \\
fio2 & 71.8 & 66.1 & 58.2 & 51.9 & 53.4 & dobutamine & 6.4 & 18.5 & 0.4 & 1.7 & 1.0 \\
ptt & 97.6 & 22.0 & 78.2 & 80.8 & 22.1 & amiodarone & 11.5 & --- & 3.2 & 6.1 & 2.1 \\
lactate & 99.8 & 91.8 & 39.5 & 45.8 & 17.9 & dexmedetomidine & 6.7 & --- & 1.9 & 11.2 & 2.2 \\
rbc & --- & --- & 98.6 & 98.4 & 92.7 & calcium\_corrected & 16.7 & --- & --- & --- & --- \\
rdw & --- & --- & 98.6 & 98.4 & 87.1 & epinephrine & 0.9 & 8.4 & 0.9 & 2.9 & 0.5 \\
calcium\_free & 99.8 & 90.6 & 49.0 & 42.6 & --- & morphine & 3.0 & 6.0 & 1.3 & 2.0 & --- \\
alt & 81.1 & 69.9 & 37.4 & 46.3 & 45.4 & bilirubin\_indirect & --- & --- & 5.8 & 6.0 & --- \\
sao2 & 97.3 & 90.8 & 36.6 & 26.2 & 28.8 & furosemide & 0.5 & --- & 3.1 & 3.5 & 0.5 \\
ast & 82.3 & 64.3 & 37.4 & 47.0 & 45.9 & dopamine & 0.1 & --- & 1.2 & 1.8 & 0.9 \\
anion\_gap & --- & --- & 98.4 & 98.6 & 75.0 & milrinone & --- & --- & 1.3 & 2.2 & 0.3 \\
bilirubin\_total & 88.2 & 53.7 & 38.4 & 46.6 & 44.7 & vasopressin & --- & 0.1 & --- & --- & 1.6 \\
\bottomrule
\end{tabular}

\end{table}

\begin{figure}[htbp]
\centering
\includegraphics[width=\textwidth]{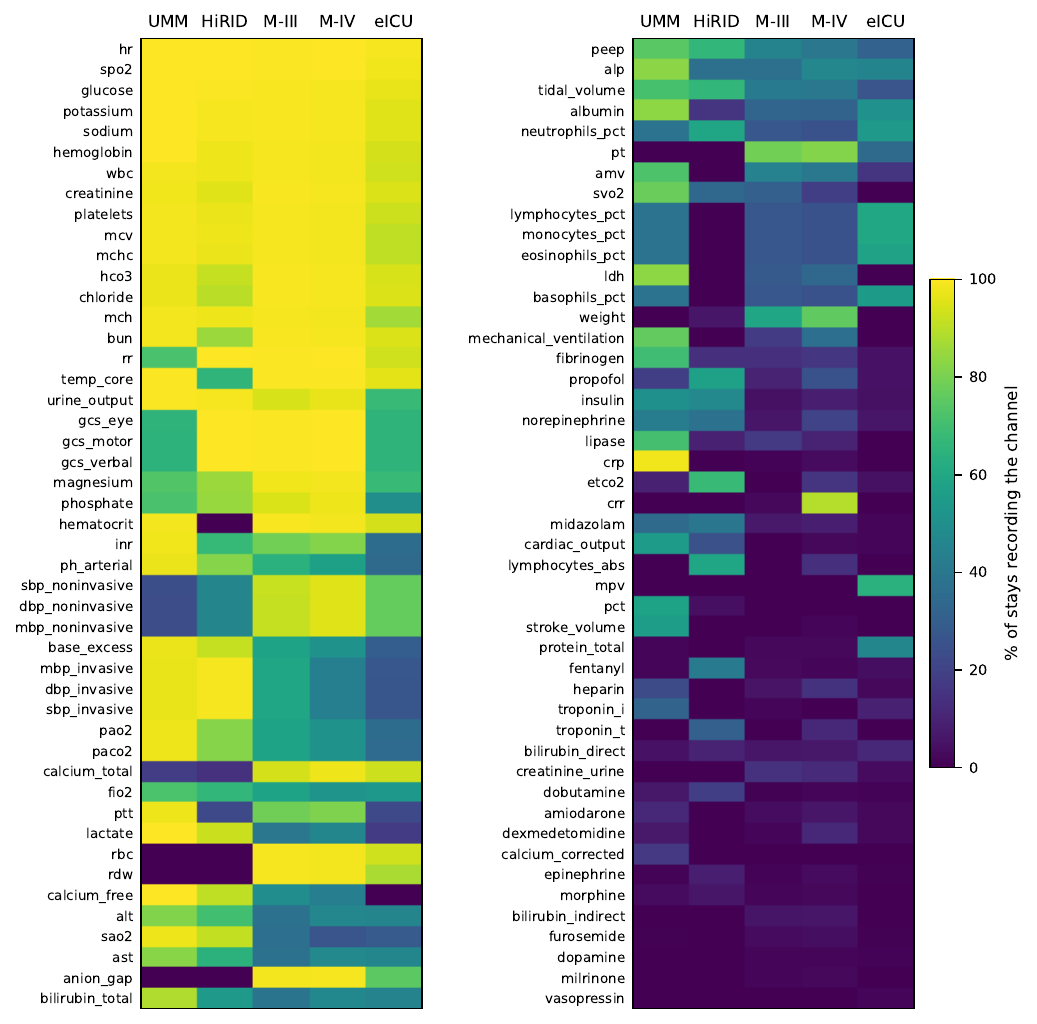}
\caption{Per-channel stay coverage. The block of channels every database records is small; most of the panel is local practice.}\label{fig:coverage}
\end{figure}

\clearpage

\section{Base pool}\label{app:models}

14 members per database, each tuned per database, selected on validation error under one
rule applied to every family, and trained on the training split alone. The two foundation models are
used \emph{zero-shot}: public checkpoints, no fitting to these databases. 
Table~\ref{tab:pool} lists the selected
configurations by family. Table~\ref{tab:manifest}
gives the same selection in full: for every (database, member) pair, the checkpoint whose
predictions were cached, its configuration, its seed, and the number of stored checkpoints in
that family on that database. The last column is why we report a selected configuration rather
than the lowest error found during search: the number of configurations differs between
families by more than an order of magnitude, so a minimum over more attempts is optimistically
biased. On \textsc{UMM}, DenseAR has 290 stored configurations against Autoformer's 17.
\textsc{locf} and \textsc{null} have no trained hyperparameters, and the two zero-shot
foundation models have no fitted configuration at all.

The three members of the extended pool (Section~\ref{sec:lgbm}) use one fixed configuration on every
database. LightGBM fits one regressor per channel with the horizon step as a feature, on seven
summaries of that channel's input window (last value, mean, minimum, maximum, count, hours since
the last observation, last minus first value) plus the static features: 200 trees, 63 leaves,
learning rate 0.05, at least 50 samples per leaf. ARIMA is a per-channel ARIMA(2,1,0) fitted by
conditional least squares on the forward-filled input window, falling back to last value plus
linear drift where the fit fails. ETS is damped Holt smoothing with fixed $\alpha=0.5$,
$\beta=0.1$ and $\phi=0.9$. Neither ARIMA nor ETS is fitted on the training split.

\begin{table}[htbp]
\centering\scriptsize
\caption{Selected configuration per architecture family and database. Names are stripped of the database, the family prefix and the run hash; \emph{default} means the family's standard configuration was selected.}\label{tab:pool}
\begin{tabular}{llllll}
\toprule
family & \textsc{UMM} & HiRID & MIMIC-III & MIMIC-IV & eICU \\
\midrule
Autoformer & ms2 & hp\_d512\_L2 & hp\_d128\_L2 & hp\_d512\_L1 & hp\_d512\_L3 \\
DLinear & ms & ms & hp\_k21\_individual & ms2 & ms2 \\
Dense DMS & hp\_d256\_L2 & ms & ms & hp\_d256\_L1 & ms \\
DenseAR & default & default & default & pteacherctrl & default \\
GRU & gru & gru & ms & ms & ms \\
Informer & hp\_d256\_L2 & hp\_d512\_L3 & hp\_d256\_L2 & hp\_d128\_L1 & hp\_d256\_L1 \\
LOCF & persistence & persistence & persistence & persistence & persistence \\
LSTM & lstm & ms & ms & ms & ms \\
Mamba & ms & m & ms & ms & m \\
Moirai & --- & --- & --- & --- & --- \\
Null & null & null & null & null & null \\
PatchTST & 4h\_d256\_L2\_lr3e4 & 4h\_d256\_L2\_lr1e3 & 4h\_d128\_L3\_lr1e3 & 4h\_d128\_L3\_lr3e4 & 4h\_d128\_L2\_lr3e4 \\
STraTS & d128\_l1\_lr1e-3\_do005 & d64\_l1\_lr5e-4\_do005 & ms & d128\_l1\_lr1e-3 & d128\_l2\_lr1e-3\_do005 \\
TiRex-2 & --- & --- & --- & --- & --- \\
\bottomrule
\end{tabular}

\end{table}

\begin{table}[htbp]
\centering\scriptsize\setlength{\tabcolsep}{2.5pt}\renewcommand{\arraystretch}{0.95}
\caption{Selected run per database and member: the checkpoint whose cached predictions enter
the pool, its width $d$, its depth (encoder/decoder where the family has both), dropout,
learning rate and seed, and \emph{ckpts}, the number of checkpoints stored for that family on
that database. Run names are stripped of the database and the family prefix; \emph{default}
means the family's standard configuration was selected.}\label{tab:manifest}
\begin{tabular}{@{}llrrrrrr@{}}
\toprule
member & run & $d$ & layers & dropout & lr & seed & ckpts \\
\midrule
\multicolumn{8}{l}{\textsc{UMM}} \\
\quad Autoformer & \texttt{ms2\_s456} & 128 & 1 & 0.1 & 0.0003 & 456 & 17 \\
\quad DLinear & \texttt{ms\_s123} & 128 & 2 & 0.2 & 0.0003 & 123 & 16 \\
\quad Dense DMS & \texttt{hp\_d256\_L2\_s123} & 256 & 2 & 0.1 & 0.0003 & 123 & 17 \\
\quad DenseAR & \texttt{hpsf\_6a82f5b0} & 512 & 1/1 & 0.0 & 0.0001 & 42 & 290 \\
\quad GRU & \texttt{hps\_b7eec8ac} & 512 & 1 & 0.1 & 0.0003 & 42 & 45 \\
\quad Informer & \texttt{hp\_d256\_L2\_s123} & 256 & 2 & 0.1 & 0.0003 & 123 & 58 \\
\quad LOCF & persistence &  &  &  &  &  & 1 \\
\quad LSTM & \texttt{hps\_01f64882} & 256 & 1 & 0.1 & 0.0003 & 42 & 45 \\
\quad Mamba & \texttt{ms\_s456} & 384 & 1 & 0.1 & 0.0003 & 456 & 44 \\
\quad PatchTST & \texttt{4h\_d256\_L2\_dr00\_lr3e4\_pl4st2\_s42} & 256 & 2 & 0.0 & 0.0003 & 42 & 100 \\
\quad STraTS & \texttt{d128\_l1\_lr1e-3\_do005} & 128 & 1 & 0.05 & 0.001 & 42 & 128 \\
\quad null & --- &  &  &  &  &  & 1 \\
\addlinespace
\multicolumn{8}{l}{\emph{HiRID}} \\
\quad Autoformer & \texttt{hp\_d512\_L2\_s123} & 512 & 2 & 0.1 & 0.0003 & 123 & 17 \\
\quad DLinear & \texttt{ms\_s123} & 128 & 2 & 0.1 & 0.0003 & 123 & 16 \\
\quad Dense DMS & \texttt{ms\_s1337} & 128 & 2 & 0.1 & 0.0003 & 1337 & 17 \\
\quad DenseAR & \texttt{hpsf\_f4e3a6c7} & 256 & 1/2 & 0.0 & 0.0003 & 42 & 280 \\
\quad GRU & \texttt{hps\_3f4077c0} & 512 & 1 & 0.1 & 0.0003 & 42 & 45 \\
\quad Informer & \texttt{hp\_d512\_L3\_s123} & 512 & 3 & 0.1 & 0.0003 & 123 & 58 \\
\quad LOCF & persistence &  &  &  &  &  & 1 \\
\quad LSTM & \texttt{ms\_s1337} & 256 & 1 & 0.1 & 0.0003 & 1337 & 45 \\
\quad Mamba & \texttt{hpsm\_2c6df0cb} & 512 & 1 & 0.1 & 0.0001 & 42 & 44 \\
\quad PatchTST & \texttt{4h\_d256\_L2\_dr00\_lr1e3\_pl4st2\_s42} & 256 & 2 & 0.0 & 0.001 & 42 & 100 \\
\quad STraTS & \texttt{d64\_l1\_lr5e-4\_do005} & 64 & 1 & 0.05 & 0.0005 & 42 & 128 \\
\quad null & --- &  &  &  &  &  & 1 \\
\addlinespace
\multicolumn{8}{l}{\emph{MIMIC-III}} \\
\quad Autoformer & \texttt{hp\_d128\_L2\_s123} & 128 & 2 & 0.1 & 0.0003 & 123 & 17 \\
\quad DLinear & \texttt{hp\_k21\_individual\_s123} & 128 & 2 & 0.1 & 0.0003 & 123 & 16 \\
\quad Dense DMS & \texttt{ms\_s123} & 128 & 2 & 0.1 & 0.0003 & 123 & 17 \\
\quad DenseAR & \texttt{s456} & 512 & 3/2 & 0.0 & 0.0001 & 456 & 273 \\
\quad GRU & \texttt{ms\_s1337} & 384 & 1 & 0.2 & 0.0003 & 1337 & 45 \\
\quad Informer & \texttt{hp\_d256\_L2\_s123} & 256 & 2 & 0.1 & 0.0003 & 123 & 58 \\
\quad LOCF & persistence &  &  &  &  &  & 1 \\
\quad LSTM & \texttt{ms\_s789} & 256 & 1 & 0.1 & 0.0003 & 789 & 45 \\
\quad Mamba & \texttt{ms\_s789} & 128 & 1 & 0.0 & 0.0003 & 789 & 44 \\
\quad PatchTST & \texttt{4h\_d128\_L3\_dr00\_lr1e3\_pl4st2\_s42} & 128 & 3 & 0.0 & 0.001 & 42 & 96 \\
\quad STraTS & \texttt{ms\_s789} & 256 & 2 & 0.1 & 0.0003 & 789 & 126 \\
\quad null & --- &  &  &  &  &  & 1 \\
\addlinespace
\multicolumn{8}{l}{\emph{MIMIC-IV}} \\
\quad Autoformer & \texttt{hp\_d512\_L1\_s123} & 512 & 1 & 0.1 & 0.0003 & 123 & 17 \\
\quad DLinear & \texttt{ms2\_s456} & 128 & 2 & 0.1 & 0.0003 & 456 & 16 \\
\quad Dense DMS & \texttt{hp\_d256\_L1\_s123} & 256 & 1 & 0.1 & 0.0003 & 123 & 17 \\
\quad DenseAR & \texttt{pteacherctrl} & 512 & 2 & 0.0 & 0.0001 & 42 & 273 \\
\quad GRU & \texttt{ms\_s123} & 512 & 1 & 0.1 & 0.0003 & 123 & 45 \\
\quad Informer & \texttt{hp\_d128\_L1\_s123} & 128 & 1 & 0.1 & 0.0003 & 123 & 58 \\
\quad LOCF & persistence &  &  &  &  &  & 1 \\
\quad LSTM & \texttt{ms\_s1337} & 384 & 1 & 0.2 & 0.0003 & 1337 & 45 \\
\quad Mamba & \texttt{ms\_s123} & 128 & 1 & 0.0 & 0.0003 & 123 & 43 \\
\quad PatchTST & \texttt{4h\_d128\_L3\_dr00\_lr3e4\_pl4st2\_s42} & 128 & 3 & 0.0 & 0.0003 & 42 & 71 \\
\quad STraTS & \texttt{d128\_l1\_lr1e-3} & 128 & 1 & 0.1 & 0.001 & 42 & 122 \\
\quad null & --- &  &  &  &  &  & 1 \\
\addlinespace
\multicolumn{8}{l}{\emph{eICU}} \\
\quad Autoformer & \texttt{hp\_d512\_L3\_s123} & 512 & 3 & 0.1 & 0.0003 & 123 & 17 \\
\quad DLinear & \texttt{ms2\_s456} & 128 & 2 & 0.1 & 0.0003 & 456 & 17 \\
\quad Dense DMS & \texttt{ms\_s456} & 128 & 2 & 0.1 & 0.0003 & 456 & 17 \\
\quad DenseAR & \texttt{hpsf\_0fc6127a} & 512 & 2 & 0.0 & 0.0001 & 42 & 258 \\
\quad GRU & \texttt{ms\_s123} & 512 & 1 & 0.1 & 0.0003 & 123 & 45 \\
\quad Informer & \texttt{hp\_d256\_L1\_s123} & 256 & 1 & 0.1 & 0.0003 & 123 & 58 \\
\quad LOCF & persistence &  &  &  &  &  & 1 \\
\quad LSTM & \texttt{ms\_s456} & 512 & 1 & 0.1 & 0.0003 & 456 & 45 \\
\quad Mamba & \texttt{hpsm\_f602f35f} & 128 & 1 & 0.0 & 0.0003 & 42 & 35 \\
\quad PatchTST & \texttt{4h\_d128\_L2\_dr00\_lr3e4\_pl4st2\_s42} & 128 & 2 & 0.0 & 0.0003 & 42 & 59 \\
\quad STraTS & \texttt{d128\_l2\_lr1e-3\_do005} & 128 & 2 & 0.05 & 0.001 & 42 & 115 \\
\quad null & --- &  &  &  &  &  & 1 \\
\bottomrule
\end{tabular}
\end{table}

\clearpage

\section{Combination algorithms}\label{app:comb}

Every combiner produces a weight tensor $w_{fhm}$ (channel, horizon step, model) and is
scored by one shared code path, so differences lie in the estimator alone. Let $\mathrm{MSE}_{fm}$
be model $m$'s masked error on channel $f$ over the fitting split, $n_f$ its observation count, and
$G_f$, $b_f$ the Gram matrix and cross-term of Section~4.1.

\begin{description}\small
\item[\normalfont\textsc{best-single}] one-hot on $\arg\min_m \overline{\mathrm{MSE}}_m$. The reference.
\item[\normalfont\textsc{mean}] $w=1/M$ everywhere. No fitting.
\item[\normalfont\textsc{g-equal-}$k$] $1/k$ on the $k$ best members \emph{pool-wide}, the same
  members on every channel.
\item[\normalfont\textsc{g-harm-}$k$] the harmonic profile $1,\tfrac12,\dots,\tfrac1k$ on those same
  pool-wide members. With \textsc{equal-}$k$ and \textsc{harm-}$k$ below it completes a
  $2\times2$ over membership and weight profile.
\item[\normalfont\textsc{equal-}$k$] $1/k$ on the $k$ best members \emph{of that channel} --- the
  matched control for \textsc{harm-}$k$, differing from it only in the weight profile.
\item[\normalfont\textsc{inv-mse}] $w_{fm}\propto 1/\mathrm{MSE}_{fm}$, per channel.
\item[\normalfont\textsc{harm-}$k$] rank the members on each channel, weight the best $k$ by
  $1,\tfrac12,\dots,\tfrac1k$ normalized. Only the ranking is estimated; the weights are fixed in
  advance, so this cannot overfit the weight values.
\item[\normalfont\textsc{select}] one-hot on $\arg\min_m \mathrm{MSE}_{fm}$, per channel, the $k=1$ end of \textsc{harm-}$k$. Fewest parameters,
  highest variance: a discrete $\arg\min$ over $M$ noisy estimates.
\item[\normalfont\textsc{ridge}] $w_f=(G_f+\lambda I)^{-1}b_f$, unconstrained; may be negative or $>1$.
\item[\normalfont\textsc{simplex}] $\min_w w^\top G_f w - 2 w^\top b_f$ subject to $w\ge 0$, $\sum_m w_m = 1$,
  by projected gradient. The constraint bounds the combination's error on that channel by that
  channel's worst member.
\item[\normalfont\textsc{g-simplex}] the same program with \emph{one} weight vector for the whole
  panel: $\min_w \sum_f (w^\top G_f w - 2 w^\top b_f + c_f)/n_f$ over the simplex, where $c_f$ is the
  sum of squared observed targets on channel $f$ and channels with $n_f=0$ are excluded. Dividing by
  $n_f$ matches the macro metric, so a densely sampled channel does not dominate the loss. This is
  the control for whether weights need to vary across channels at all.
\item[\normalfont\textsc{bma}] Bayesian model averaging, $w_{fm}\propto\exp(-\tfrac{N_f}{2}\mathrm{MSE}_{fm}/\sigma_f^2)$
  with $\sigma_f^2$ the best member's error on the channel and flat priors. $N_f$ is the observed-cell
  count; \textsc{bma}$_{\mathrm{stays}}$ uses the stay count instead, since cells within a stay are
  not independent.
\item[\normalfont\textsc{hedge}] exponentiated-gradient weights $w_{fm}\propto\exp(-\eta_f T_f\,
  \mathrm{MSE}_{fm}/\bar{\mathrm{MSE}}_f)$ with the standard $\eta_f=\sqrt{8\ln M/T_f}$, so the
  exponent grows as $\sqrt{T_f}$ rather than $T_f$.
\item[\normalfont\textsc{greedy}] forward selection with replacement: repeatedly add the member that most
  reduces error, then average the selections.
\item[\normalfont\textsc{simplex-shrunk}$\kappa$] $w_f = \alpha_f w^{\text{simplex}}_f + (1-\alpha_f) w^{\text{prior}}_f$
  with $\alpha_f = n_f/(n_f+\kappa)$ and $w^{\text{prior}}$ the \textsc{harm-5} ranking weights for
  that channel, so a dense channel keeps its fitted solution and a thin one falls back to the
  ranking prior; a channel with no support at all falls back to one hour-pooled weight vector
  estimated across every channel. $\kappa\in\{100,500\}$; $\kappa=500$ is \textsc{ensemble}.
\end{description}

\textsc{select} and \textsc{ridge} additionally have hour-resolved counterparts, \textsc{sel-}$h$ and ridge-$h$, in which $w$ varies with $h$.

\clearpage

\section{Combiner ladder: validation and test}\label{app:ladder}

\textsc{Ensemble} (\textsc{simplex-shrunk}500) was chosen on the development column of
Table~\ref{tab:ladder}, where it has the best mean; test was opened
afterwards. Most fitted combiners improve from development to test, where the final fit sees twice
the stays, and the ordering changes with it: the shrunk and unshrunk simplex swap places
(Section~\ref{sec:findings}). Unconstrained ridge is the exception on HiRID, moving from
$-3.59$\,\% to $+1.77$\,\%. The two protocols also differ in scoring population, not only in
fitting size.

\begin{table}[htbp]
\centering\small
\caption{Combiner ladder, as percentage change against the best single model (negative is better). \emph{Dev} fits on a stay-grouped half of validation and scores on the disjoint half; \emph{test} fits on all of validation and scores on test, so the pair differs both in how much data the weights are fitted on and in the population they are scored on. Rows are ordered by the test mean. \textsc{Ensemble} is \textsc{simplex-shrunk}500, chosen on the \emph{dev} columns before test was opened.}\label{tab:ladder}
\setlength{\tabcolsep}{3pt}
\begin{tabular}{lrrrrrrrrrrrr}
\toprule
 & \multicolumn{2}{c}{\textsc{UMM}} & \multicolumn{2}{c}{HiRID} & \multicolumn{2}{c}{MIMIC-III} & \multicolumn{2}{c}{MIMIC-IV} & \multicolumn{2}{c}{eICU} & \multicolumn{2}{c}{mean} \\
\cmidrule(lr){2-3}\cmidrule(lr){4-5}\cmidrule(lr){6-7}\cmidrule(lr){8-9}\cmidrule(lr){10-11}\cmidrule(lr){12-13}
combiner & dev & test & dev & test & dev & test & dev & test & dev & test & dev & test \\
\midrule
\textsc{greedy} & -8.85 & -10.69 & -3.21 & -12.40 & -5.39 & -9.87 & -5.57 & -6.57 & -4.89 & -6.34 & -5.58 & -9.17 \\
simplex & -10.04 & -10.69 & -6.80 & -12.42 & -5.33 & -9.83 & -5.49 & -6.54 & -4.96 & -6.30 & -6.52 & -9.16 \\
shrunk100 & -10.79 & -10.94 & -7.97 & -10.29 & -5.47 & -9.83 & -5.52 & -6.54 & -4.97 & -6.10 & -6.94 & -8.74 \\
\textsc{ens.} & -10.67 & -10.95 & -8.00 & -10.01 & -5.61 & -9.79 & -5.57 & -6.51 & -5.00 & -6.09 & -6.97 & -8.67 \\
\textsc{harm-5} & -8.02 & -8.79 & -4.01 & -8.52 & -5.12 & -8.48 & -4.70 & -4.97 & -4.12 & -5.01 & -5.19 & -7.15 \\
g-simplex & -10.83 & -9.95 & -6.78 & -7.91 & -6.24 & -7.51 & -5.18 & -5.06 & -4.94 & -5.02 & -6.79 & -7.09 \\
equal-5 & -8.67 & -8.98 & -3.74 & -7.36 & -5.05 & -8.85 & -4.08 & -5.05 & -3.75 & -4.62 & -5.06 & -6.97 \\
\textsc{harm-3} & -7.13 & -8.10 & -3.17 & -8.08 & -4.53 & -7.57 & -4.41 & -4.14 & -3.87 & -4.46 & -4.62 & -6.47 \\
equal-3 & -7.93 & -8.58 & -3.05 & -6.96 & -4.88 & -7.95 & -4.37 & -3.99 & -3.79 & -4.21 & -4.80 & -6.34 \\
inv-MSE & -9.74 & -9.18 & -5.63 & -7.80 & -4.69 & -7.34 & -2.81 & -3.39 & -3.72 & -3.67 & -5.32 & -6.28 \\
ridge & -3.57 & -7.77 & -3.59 & 1.77 & 14.49 & -7.86 & -4.83 & -6.39 & -3.35 & -4.30 & -0.17 & -4.91 \\
g-equal-5 & -7.30 & -8.14 & -5.57 & -6.36 & -4.14 & -4.35 & -2.97 & -2.72 & -1.92 & -2.87 & -4.38 & -4.89 \\
mean & -8.10 & -8.01 & -4.50 & -6.26 & -3.63 & -5.86 & -1.20 & -1.07 & -3.14 & -2.67 & -4.11 & -4.77 \\
g-harm-5 & -7.11 & -7.44 & -4.68 & -4.57 & -3.95 & -3.81 & -3.21 & -3.04 & -1.91 & -2.80 & -4.17 & -4.33 \\
BMA$_{\mathrm{stays}}$ & -4.72 & -5.28 & -4.65 & -7.91 & -0.73 & -3.93 & -1.17 & -1.57 & -1.32 & -1.57 & -2.52 & -4.05 \\
\textsc{sel-}$h$ & -2.32 & -4.64 & 0.33 & -7.26 & -0.81 & -5.34 & -1.06 & -0.80 & -1.61 & -1.81 & -1.09 & -3.97 \\
Hedge & -5.37 & -5.51 & -4.71 & -5.38 & -1.62 & -4.25 & -1.62 & -1.80 & -1.32 & -2.11 & -2.93 & -3.81 \\
\textsc{select} & -2.45 & -4.89 & -0.64 & -7.30 & -0.56 & -3.84 & -1.04 & -1.48 & -1.45 & -1.52 & -1.23 & -3.81 \\
g-equal-3 & -6.04 & -8.05 & -4.02 & -2.34 & -3.72 & -3.08 & -2.68 & -2.68 & -1.66 & -2.47 & -3.62 & -3.72 \\
ridge-$h$ & -2.73 & -6.81 & -2.69 & 3.33 & 15.66 & -7.27 & -2.80 & -4.47 & -2.23 & -3.38 & 1.04 & -3.72 \\
g-harm-3 & -6.61 & -6.68 & -3.49 & -2.15 & -3.36 & -2.84 & -2.72 & -2.71 & -1.58 & -2.39 & -3.55 & -3.35 \\
BMA & -4.93 & -4.98 & -4.56 & -4.74 & -1.33 & -3.95 & -1.07 & -1.49 & -1.38 & -1.54 & -2.65 & -3.34 \\
\textsc{best} & 0.00 & 0.00 & 0.00 & 0.00 & 0.00 & 0.00 & 0.00 & 0.00 & 0.00 & 0.00 & 0.00 & 0.00 \\
\bottomrule
\end{tabular}
\end{table}

\clearpage

\section{Uncertainty}\label{app:unc}

Paired approximate randomization test over stays (10\,000 draws), each combiner against
the validation-chosen best single model. Because MSE-per-var is a ratio of sums per channel and then
a mean over channels --- not a mean over units --- the full statistic is recomputed on every draw
rather than approximated by resampling per-stay scores. $p=10^{-4}$ is the floor, $1/(R+1)$. Results are in Table~\ref{tab:unc}.

\begin{table}[htbp]
\centering\small
\caption{Change against the best single model on test, with 95\% bootstrap confidence intervals over stays. $\dagger$ marks $p \geq 0.05$ and $^{*}$ marks $10^{-4} < p < 0.05$ under a paired randomization test.}\label{tab:unc}
\begin{tabular}{lrrrrr}
\toprule
combiner & \textsc{UMM} & HiRID & MIMIC-III & MIMIC-IV & eICU \\
\midrule
\textsc{ens.} & -10.9 {\tiny [-13.3, -8.2]} & -10.0 {\tiny [-15.5, -6.9]} & -9.8 {\tiny [-11.5, -7.9]} & -6.5 {\tiny [-8.6, -4.9]} & -6.1 {\tiny [-6.7, -5.2]} \\
simplex & -10.7 {\tiny [-13.1, -8.0]} & -12.4 {\tiny [-16.7, -8.9]} & -9.8 {\tiny [-11.8, -7.6]} & -6.5 {\tiny [-8.6, -4.9]} & -6.3 {\tiny [-6.9, -5.3]} \\
\textsc{greedy} & -10.7 {\tiny [-13.1, -7.9]} & -12.4 {\tiny [-16.7, -8.8]} & -9.9 {\tiny [-11.8, -7.6]} & -6.6 {\tiny [-8.6, -5.0]} & -6.3 {\tiny [-6.9, -5.3]} \\
\textsc{harm-5} & -8.8 {\tiny [-11.2, -6.0]} & -8.5 {\tiny [-14.2, -5.3]} & -8.5 {\tiny [-10.1, -6.8]} & -5.0 {\tiny [-6.3, -3.8]} & -5.0 {\tiny [-5.8, -3.8]} \\
\textsc{harm-3} & -8.1 {\tiny [-10.4, -5.4]} & -8.1 {\tiny [-13.7, -4.8]} & -7.6 {\tiny [-9.2, -6.0]} & -4.1 {\tiny [-5.2, -3.2]} & -4.5 {\tiny [-5.2, -3.2]} \\
\textsc{select} & -4.9 {\tiny [-9.3, -0.6]}$^{*}$ & -7.3 {\tiny [-12.3, -3.3]} & -3.8 {\tiny [-5.5, -2.2]} & -1.5 {\tiny [-2.3, -0.7]}$^{*}$ & -1.5 {\tiny [-2.5, -0.0]}$^{*}$ \\
mean & -8.0 {\tiny [-10.7, -4.7]} & -6.3 {\tiny [-8.7, -4.0]} & -5.9 {\tiny [-8.0, -3.6]} & -1.1 {\tiny [-3.9, 1.5]}$^{\dagger}$ & -2.7 {\tiny [-3.2, -1.6]} \\
\bottomrule
\end{tabular}

\end{table}

$^{\dagger}$ not significant at $p<0.05$. Every soft combiner, fitted or rank-weighted, is
significant at the floor on every database. Per-channel \textsc{select} is significant everywhere,
but above the floor on \textsc{UMM} ($p=0.04$), MIMIC-IV ($p=4\times10^{-4}$) and eICU ($p=0.04$); the only failure is the unweighted mean on MIMIC-IV,
i.e.\ naive averaging is the one combiner that can fail to beat a single model. Brackets give a
95\% percentile confidence interval from 2\,000 bootstrap resamples of stays. Table~\ref{tab:lgbm} resamples patients instead.

\clearpage

\section{Which model wins which cells}\label{app:win}

Share of well-supported (variable, hour) test cells on which each family has the lowest
error ($\geq 200$ observations on the scoring split), in Table~\ref{tab:win}. Shares sum to
100\% per column. Table~\ref{tab:win} aggregates over the horizon; Figure~\ref{fig:maps} shows the
same quantity resolved by (variable, hour) on one database.

\begin{figure}[htbp]
\centering
\includegraphics[width=\textwidth]{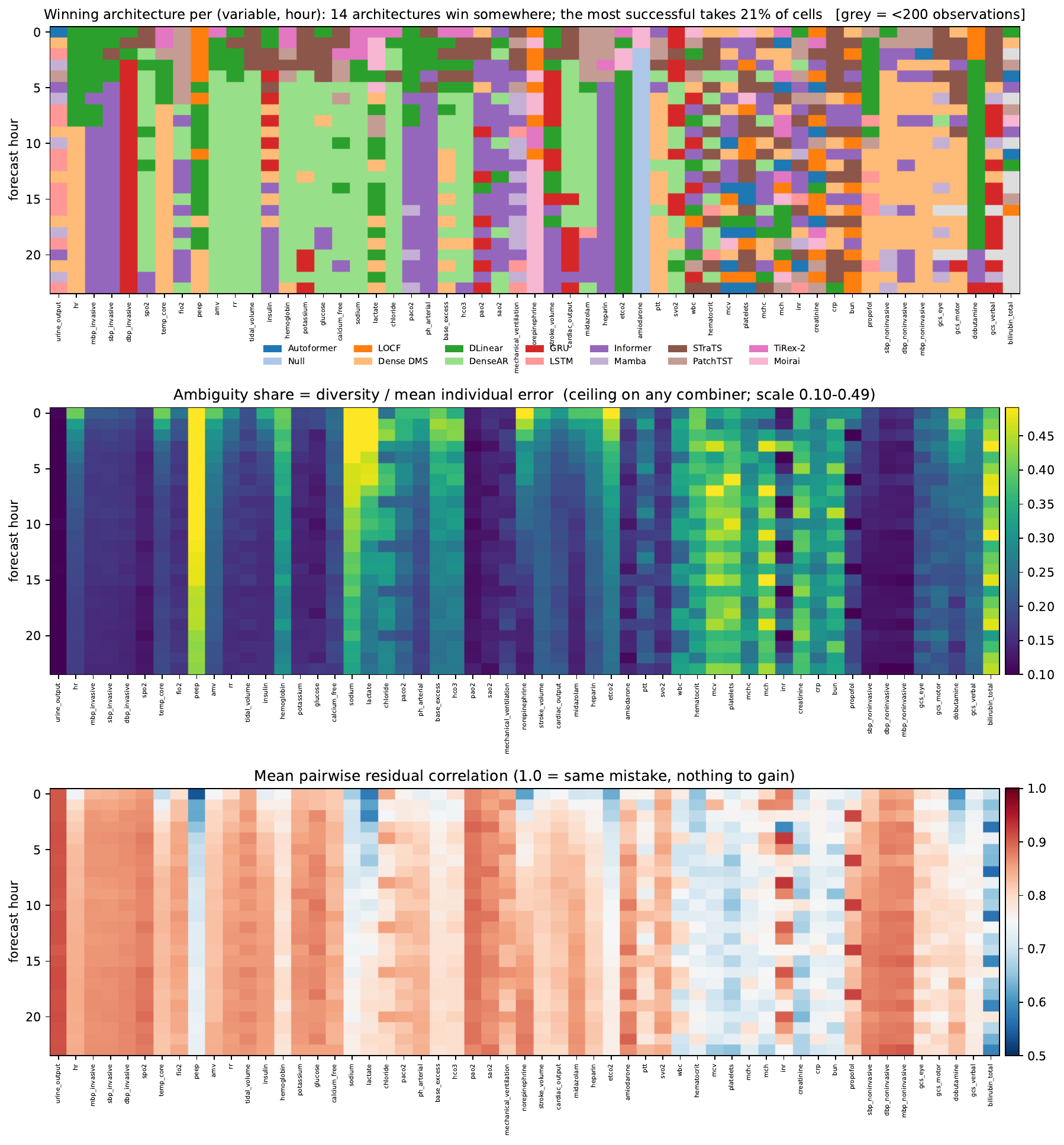}
\caption{Per-(variable, hour) structure on \textsc{UMM} validation, the densest database and so the
one where most channels clear the support filter. \textbf{Top:} which architecture wins each cell.
\textbf{Center:} ambiguity share, the weighted spread of the members about their own mean
under equal weights; it bounds the gain of the equal-weight average only, not of a fitted combiner. \textbf{Bottom:}
mean pairwise residual correlation. Cells with fewer than 200 observations in \emph{either} the
fitting or the scoring half are grayed, so a winner picked from a handful of observations cannot be
over-read. No architecture owns the grid: the winner changes across variables and, less often,
along the horizon.}\label{fig:maps}
\end{figure}

\begin{table}[htbp]
\centering\small
\caption{Share of well-supported (variable, hour) test cells won by each family (\%).}\label{tab:win}
\begin{tabular}{lrrrrrr}
\toprule
family & \textsc{UMM} & HiRID & MIMIC-III & MIMIC-IV & eICU & mean \\
\midrule
STraTS & 12.2 & 24.4 & 30.0 & 40.0 & 35.5 & 28.4 \\
DenseAR & 22.8 & 21.1 & 34.2 & 20.6 & 35.8 & 26.9 \\
Informer & 21.2 & 9.3 & 7.1 & 7.5 & 2.9 & 9.6 \\
GRU & 6.2 & 11.1 & 2.6 & 14.1 & 3.9 & 7.6 \\
Dense DMS & 9.1 & 8.7 & 7.9 & 3.0 & 7.3 & 7.2 \\
DLinear & 13.9 & 16.5 & 1.7 & 1.1 & 0.8 & 6.8 \\
PatchTST & 5.7 & 2.9 & 7.1 & 5.2 & 5.6 & 5.3 \\
LOCF & 3.3 & 2.5 & 3.4 & 2.7 & 3.0 & 3.0 \\
LSTM & 1.7 & 0.5 & 0.4 & 0.9 & 2.7 & 1.3 \\
Moirai & 1.3 & 1.4 & 1.8 & 1.3 & 0.5 & 1.2 \\
TiRex-2 & 1.9 & 0.9 & 0.7 & 1.0 & 1.0 & 1.1 \\
Null & 0.1 & 0.0 & 2.8 & 1.8 & 0.1 & 1.0 \\
Autoformer & 0.3 & 0.6 & 0.3 & 0.1 & 0.7 & 0.4 \\
Mamba & 0.3 & 0.1 & 0.0 & 0.7 & 0.2 & 0.3 \\
\bottomrule
\end{tabular}

\end{table}

\clearpage

\section{Pool diversity}\label{app:div}

Mean pairwise residual correlation between pool members, averaged over the five
databases and weighted by observation count, in Table~\ref{tab:div}. What a combiner can exploit is
exactly what this matrix is missing from one: two members correlating at $0.95$ leave almost nothing
to arbitrate. Three structures are visible. The seven in-domain sequence models form a block at
$0.91$--$0.95$, so they are close to interchangeable as ensemble members. The trivial baselines are
the most distant members of all --- \textsc{null} at $0.42$--$0.68$ and \textsc{locf} at
$0.77$--$0.90$ --- which is why they still earn weight. And the two zero-shot foundation models correlate most with PatchTST ($0.92$--$0.93$) and with
each other ($0.91$), and least with the in-domain sequence models ($0.81$--$0.84$), which makes them
the least correlated learned members.

\begin{table}[htbp]
\centering\scriptsize
\caption{Mean pairwise residual correlation between pool members, averaged over the five databases; $1.00$ on the diagonal by construction, lower is more complementary.}\label{tab:div}
\providecommand{\rot}[1]{\rotatebox[origin=l]{60}{#1}}
\begin{tabular}{lrrrrrrrrrrrrrr}
\toprule
 & \rot{DenseAR} & \rot{Informer} & \rot{DenseDMS} & \rot{GRU} & \rot{LSTM} & \rot{STraTS} & \rot{Mamba} & \rot{DLinear} & \rot{PatchTST} & \rot{Autoformer} & \rot{Moirai} & \rot{TiRex-2} & \rot{LOCF} & \rot{null} \\
\midrule
DenseAR & 1.00 & 0.95 & 0.94 & 0.95 & 0.94 & 0.92 & 0.94 & 0.90 & 0.90 & 0.86 & 0.83 & 0.84 & 0.79 & 0.67 \\
Informer & 0.95 & 1.00 & 0.94 & 0.95 & 0.94 & 0.92 & 0.94 & 0.90 & 0.90 & 0.87 & 0.83 & 0.84 & 0.79 & 0.66 \\
DenseDMS & 0.94 & 0.94 & 1.00 & 0.94 & 0.94 & 0.91 & 0.94 & 0.89 & 0.89 & 0.86 & 0.82 & 0.84 & 0.78 & 0.67 \\
GRU & 0.95 & 0.95 & 0.94 & 1.00 & 0.95 & 0.92 & 0.95 & 0.90 & 0.90 & 0.86 & 0.83 & 0.84 & 0.80 & 0.67 \\
LSTM & 0.94 & 0.94 & 0.94 & 0.95 & 1.00 & 0.92 & 0.94 & 0.89 & 0.89 & 0.85 & 0.82 & 0.83 & 0.78 & 0.68 \\
STraTS & 0.92 & 0.92 & 0.91 & 0.92 & 0.92 & 1.00 & 0.92 & 0.87 & 0.87 & 0.85 & 0.81 & 0.82 & 0.77 & 0.64 \\
Mamba & 0.94 & 0.94 & 0.94 & 0.95 & 0.94 & 0.92 & 1.00 & 0.90 & 0.89 & 0.86 & 0.83 & 0.84 & 0.78 & 0.67 \\
DLinear & 0.90 & 0.90 & 0.89 & 0.90 & 0.89 & 0.87 & 0.90 & 1.00 & 0.93 & 0.89 & 0.87 & 0.87 & 0.81 & 0.65 \\
PatchTST & 0.90 & 0.90 & 0.89 & 0.90 & 0.89 & 0.87 & 0.89 & 0.93 & 1.00 & 0.92 & 0.92 & 0.93 & 0.88 & 0.57 \\
Autoformer & 0.86 & 0.87 & 0.86 & 0.86 & 0.85 & 0.85 & 0.86 & 0.89 & 0.92 & 1.00 & 0.83 & 0.87 & 0.77 & 0.55 \\
Moirai & 0.83 & 0.83 & 0.82 & 0.83 & 0.82 & 0.81 & 0.83 & 0.87 & 0.92 & 0.83 & 1.00 & 0.91 & 0.90 & 0.48 \\
TiRex-2 & 0.84 & 0.84 & 0.84 & 0.84 & 0.83 & 0.82 & 0.84 & 0.87 & 0.93 & 0.87 & 0.91 & 1.00 & 0.85 & 0.48 \\
LOCF & 0.79 & 0.79 & 0.78 & 0.80 & 0.78 & 0.77 & 0.78 & 0.81 & 0.88 & 0.77 & 0.90 & 0.85 & 1.00 & 0.42 \\
null & 0.67 & 0.66 & 0.67 & 0.67 & 0.68 & 0.64 & 0.67 & 0.65 & 0.57 & 0.55 & 0.48 & 0.48 & 0.42 & 1.00 \\
\bottomrule
\end{tabular}

\end{table}

\clearpage

\section{Results by channel group}\label{app:system}

MSE-per-var on test for each pool family and for the combiner chosen on validation
(\textsc{ensemble}), within each channel group. Channels carrying fewer than 100 observations are
excluded, as  in every group-level table here. One table per database follows
(Tables~\ref{tab:sysumm}--\ref{tab:syseicu}). \textbf{The combination  beats every pool member in all 50
group\,$\times$\,database cells}, ten groups of which the tables show seven (\emph{freq:mid},
\emph{type:ventilation} and the channels outside \emph{sofa\_relevant} are omitted). Without the support filter it loses two of them --- eICU
ventilation and HiRID medications --- but both losses are produced entirely by channels too thin to
estimate a per-channel weight on, which is the artifact the filter exists to remove.

\begin{table}[htbp]
\centering\footnotesize
\caption{\textbf{UMM:} MSE-per-var by channel group (test). \textsc{ensemble} is the combiner chosen on validation.}\label{tab:sysumm}
\begin{tabular}{lrrrrrrr}
\toprule
family & all\_charted & type:lab & type:vital & type:medication & sofa\_relevant & freq:low & freq:high \\
\midrule
ENSEMBLE & 0.530 & 0.469 & 0.444 & 0.887 & 0.844 & 0.831 & 0.407 \\
\textsc{greedy} & 0.532 & 0.469 & 0.444 & 0.901 & 0.856 & 0.838 & 0.407 \\
harm-5 & 0.543 & 0.484 & 0.457 & 0.897 & 0.856 & 0.851 & 0.412 \\
mean & 0.549 & 0.498 & 0.455 & 0.882 & 0.870 & 0.868 & 0.414 \\
select & 0.567 & 0.505 & 0.475 & 0.945 & 0.883 & 0.898 & 0.420 \\
Informer & 0.577 & 0.501 & 0.471 & 1.021 & 0.898 & 0.923 & 0.425 \\
Dense DMS & 0.580 & 0.507 & 0.481 & 1.008 & 0.890 & 0.923 & 0.428 \\
GRU & 0.591 & 0.551 & 0.481 & 0.915 & 0.876 & 0.938 & 0.429 \\
DenseAR & 0.596 & 0.533 & 0.477 & 1.017 & 0.956 & 0.972 & 0.426 \\
STraTS & 0.601 & 0.526 & 0.496 & 1.031 & 0.965 & 0.942 & 0.453 \\
DLinear & 0.605 & 0.571 & 0.504 & 0.902 & 0.918 & 0.972 & 0.434 \\
PatchTST & 0.617 & 0.580 & 0.518 & 0.919 & 0.928 & 0.972 & 0.457 \\
LSTM & 0.649 & 0.610 & 0.494 & 1.053 & 1.024 & 1.064 & 0.443 \\
Autoformer & 0.661 & 0.619 & 0.552 & 1.003 & 0.981 & 1.032 & 0.502 \\
Moirai & 0.662 & 0.622 & 0.588 & 0.942 & 0.954 & 1.011 & 0.526 \\
Mamba & 0.664 & 0.633 & 0.506 & 1.050 & 1.030 & 1.068 & 0.442 \\
TiRex-2 & 0.667 & 0.628 & 0.578 & 0.975 & 0.957 & 1.021 & 0.515 \\
LOCF & 0.702 & 0.638 & 0.646 & 1.000 & 1.040 & 1.022 & 0.603 \\
Null & 1.060 & 1.048 & 0.981 & 1.256 & 1.324 & 1.260 & 0.974 \\
\bottomrule
\end{tabular}

\end{table}

\begin{table}[htbp]
\centering\footnotesize
\caption{\textbf{HiRID:} MSE-per-var by channel group (test). \textsc{ensemble} is the combiner chosen on validation.}\label{tab:syshirid}
\begin{tabular}{lrrrrrrr}
\toprule
family & all\_charted & type:lab & type:vital & type:medication & sofa\_relevant & freq:low & freq:high \\
\midrule
ENSEMBLE & 0.429 & 0.346 & 0.393 & 0.865 & 0.545 & 0.514 & 0.425 \\
\textsc{greedy} & 0.430 & 0.347 & 0.393 & 0.866 & 0.546 & 0.515 & 0.425 \\
harm-5 & 0.436 & 0.355 & 0.397 & 0.871 & 0.550 & 0.526 & 0.428 \\
select & 0.455 & 0.373 & 0.407 & 0.916 & 0.562 & 0.553 & 0.441 \\
mean & 0.457 & 0.366 & 0.422 & 0.937 & 0.595 & 0.570 & 0.437 \\
DenseAR & 0.487 & 0.395 & 0.417 & 1.050 & 0.657 & 0.637 & 0.445 \\
Informer & 0.488 & 0.380 & 0.442 & 1.079 & 0.670 & 0.627 & 0.453 \\
GRU & 0.491 & 0.397 & 0.428 & 1.049 & 0.670 & 0.644 & 0.447 \\
DLinear & 0.495 & 0.410 & 0.462 & 0.945 & 0.597 & 0.631 & 0.459 \\
Dense DMS & 0.497 & 0.391 & 0.428 & 1.125 & 0.703 & 0.634 & 0.465 \\
PatchTST & 0.513 & 0.424 & 0.483 & 0.965 & 0.607 & 0.628 & 0.492 \\
STraTS & 0.540 & 0.456 & 0.473 & 1.082 & 0.690 & 0.786 & 0.446 \\
LSTM & 0.547 & 0.463 & 0.464 & 1.123 & 0.741 & 0.779 & 0.456 \\
Autoformer & 0.559 & 0.469 & 0.522 & 1.036 & 0.647 & 0.675 & 0.537 \\
Moirai & 0.561 & 0.457 & 0.553 & 1.032 & 0.646 & 0.651 & 0.570 \\
Mamba & 0.568 & 0.486 & 0.488 & 1.140 & 0.747 & 0.826 & 0.459 \\
TiRex-2 & 0.581 & 0.481 & 0.554 & 1.089 & 0.683 & 0.666 & 0.566 \\
LOCF & 0.591 & 0.468 & 0.585 & 1.113 & 0.684 & 0.649 & 0.635 \\
Null & 1.010 & 0.951 & 0.989 & 1.328 & 1.299 & 1.195 & 0.930 \\
\bottomrule
\end{tabular}

\end{table}

\begin{table}[htbp]
\centering\footnotesize
\caption{\textbf{MIMIC-III:} MSE-per-var by channel group (test). \textsc{ensemble} is the combiner chosen on validation.}\label{tab:sysmimic-iii}
\begin{tabular}{lrrrrrrr}
\toprule
family & all\_charted & type:lab & type:vital & type:medication & sofa\_relevant & freq:low & freq:high \\
\midrule
\textsc{greedy} & 0.403 & 0.394 & 0.451 & 0.391 & 0.352 & 0.459 & 0.449 \\
ENSEMBLE & 0.404 & 0.394 & 0.452 & 0.390 & 0.352 & 0.459 & 0.449 \\
harm-5 & 0.410 & 0.400 & 0.453 & 0.402 & 0.363 & 0.469 & 0.452 \\
mean & 0.424 & 0.421 & 0.468 & 0.397 & 0.365 & 0.492 & 0.463 \\
select & 0.429 & 0.423 & 0.459 & 0.433 & 0.385 & 0.504 & 0.458 \\
Informer & 0.444 & 0.433 & 0.466 & 0.474 & 0.406 & 0.534 & 0.465 \\
DenseAR & 0.447 & 0.440 & 0.459 & 0.483 & 0.408 & 0.560 & 0.458 \\
GRU & 0.450 & 0.447 & 0.466 & 0.465 & 0.397 & 0.550 & 0.466 \\
Dense DMS & 0.455 & 0.446 & 0.484 & 0.468 & 0.430 & 0.534 & 0.484 \\
STraTS & 0.471 & 0.446 & 0.471 & 0.581 & 0.439 & 0.598 & 0.468 \\
LSTM & 0.473 & 0.480 & 0.473 & 0.473 & 0.424 & 0.593 & 0.474 \\
Mamba & 0.485 & 0.486 & 0.473 & 0.526 & 0.451 & 0.630 & 0.472 \\
PatchTST & 0.487 & 0.496 & 0.527 & 0.431 & 0.402 & 0.578 & 0.515 \\
DLinear & 0.510 & 0.519 & 0.549 & 0.455 & 0.428 & 0.597 & 0.530 \\
Moirai & 0.534 & 0.535 & 0.615 & 0.455 & 0.434 & 0.604 & 0.596 \\
Autoformer & 0.536 & 0.543 & 0.552 & 0.524 & 0.464 & 0.648 & 0.555 \\
LOCF & 0.562 & 0.540 & 0.707 & 0.475 & 0.465 & 0.598 & 0.666 \\
TiRex-2 & 0.569 & 0.561 & 0.578 & 0.608 & 0.561 & 0.691 & 0.595 \\
Null & 0.933 & 0.981 & 0.967 & 0.774 & 1.045 & 0.946 & 0.938 \\
\bottomrule
\end{tabular}

\end{table}

\begin{table}[htbp]
\centering\footnotesize
\caption{\textbf{MIMIC-IV:} MSE-per-var by channel group (test). \textsc{ensemble} is the combiner chosen on validation.}\label{tab:sysmimic-iv}
\begin{tabular}{lrrrrrrr}
\toprule
family & all\_charted & type:lab & type:vital & type:medication & sofa\_relevant & freq:low & freq:high \\
\midrule
\textsc{greedy} & 0.414 & 0.389 & 0.507 & 0.397 & 0.292 & 0.505 & 0.429 \\
ENSEMBLE & 0.414 & 0.389 & 0.507 & 0.396 & 0.292 & 0.505 & 0.429 \\
harm-5 & 0.421 & 0.399 & 0.511 & 0.400 & 0.292 & 0.521 & 0.430 \\
select & 0.437 & 0.415 & 0.523 & 0.418 & 0.305 & 0.551 & 0.438 \\
mean & 0.438 & 0.417 & 0.519 & 0.426 & 0.301 & 0.534 & 0.443 \\
DenseAR & 0.443 & 0.422 & 0.530 & 0.425 & 0.315 & 0.553 & 0.446 \\
GRU & 0.448 & 0.434 & 0.520 & 0.422 & 0.309 & 0.571 & 0.441 \\
Informer & 0.449 & 0.431 & 0.529 & 0.431 & 0.311 & 0.573 & 0.443 \\
Dense DMS & 0.453 & 0.437 & 0.531 & 0.430 & 0.319 & 0.581 & 0.444 \\
LSTM & 0.457 & 0.444 & 0.525 & 0.437 & 0.320 & 0.583 & 0.448 \\
STraTS & 0.470 & 0.447 & 0.532 & 0.502 & 0.337 & 0.640 & 0.442 \\
Mamba & 0.497 & 0.503 & 0.538 & 0.459 & 0.335 & 0.699 & 0.448 \\
PatchTST & 0.530 & 0.498 & 0.583 & 0.615 & 0.344 & 0.632 & 0.497 \\
DLinear & 0.533 & 0.530 & 0.600 & 0.484 & 0.369 & 0.662 & 0.521 \\
Autoformer & 0.554 & 0.526 & 0.600 & 0.635 & 0.379 & 0.678 & 0.511 \\
Moirai & 0.589 & 0.528 & 0.677 & 0.726 & 0.382 & 0.657 & 0.570 \\
TiRex-2 & 0.608 & 0.560 & 0.638 & 0.785 & 0.396 & 0.674 & 0.555 \\
LOCF & 0.624 & 0.534 & 0.772 & 0.775 & 0.413 & 0.660 & 0.667 \\
Null & 0.986 & 1.042 & 1.010 & 0.821 & 0.844 & 1.091 & 0.943 \\
\bottomrule
\end{tabular}

\end{table}

\begin{table}[htbp]
\centering\footnotesize
\caption{\textbf{eICU:} MSE-per-var by channel group (test). \textsc{ensemble} is the combiner chosen on validation.}\label{tab:syseicu}
\begin{tabular}{lrrrrrrr}
\toprule
family & all\_charted & type:lab & type:vital & type:medication & sofa\_relevant & freq:low & freq:high \\
\midrule
\textsc{greedy} & 0.394 & 0.365 & 0.461 & 0.421 & 0.336 & 0.433 & 0.385 \\
ENSEMBLE & 0.394 & 0.365 & 0.461 & 0.421 & 0.335 & 0.433 & 0.386 \\
harm-5 & 0.400 & 0.371 & 0.464 & 0.430 & 0.339 & 0.444 & 0.388 \\
mean & 0.410 & 0.385 & 0.471 & 0.426 & 0.346 & 0.451 & 0.400 \\
select & 0.414 & 0.381 & 0.473 & 0.467 & 0.357 & 0.471 & 0.395 \\
DenseAR & 0.418 & 0.388 & 0.473 & 0.469 & 0.367 & 0.480 & 0.396 \\
Informer & 0.422 & 0.393 & 0.479 & 0.465 & 0.367 & 0.476 & 0.403 \\
GRU & 0.425 & 0.397 & 0.476 & 0.468 & 0.374 & 0.484 & 0.402 \\
LSTM & 0.427 & 0.401 & 0.477 & 0.466 & 0.368 & 0.483 & 0.405 \\
Dense DMS & 0.435 & 0.407 & 0.491 & 0.469 & 0.389 & 0.482 & 0.416 \\
STraTS & 0.447 & 0.388 & 0.488 & 0.608 & 0.442 & 0.544 & 0.401 \\
Mamba & 0.460 & 0.433 & 0.489 & 0.534 & 0.397 & 0.557 & 0.411 \\
PatchTST & 0.462 & 0.442 & 0.520 & 0.472 & 0.388 & 0.520 & 0.444 \\
DLinear & 0.483 & 0.468 & 0.538 & 0.481 & 0.410 & 0.534 & 0.467 \\
Autoformer & 0.506 & 0.474 & 0.547 & 0.576 & 0.436 & 0.590 & 0.472 \\
Moirai & 0.506 & 0.475 & 0.604 & 0.506 & 0.419 & 0.560 & 0.508 \\
LOCF & 0.526 & 0.480 & 0.661 & 0.522 & 0.430 & 0.566 & 0.550 \\
TiRex-2 & 0.531 & 0.502 & 0.575 & 0.588 & 0.464 & 0.608 & 0.504 \\
Null & 0.995 & 0.996 & 1.036 & 0.955 & 0.992 & 1.007 & 0.992 \\
\bottomrule
\end{tabular}

\end{table}

\clearpage

\section{Gain by horizon}\label{app:horizon}

Combination gain of the combiner chosen on validation against the best single model, by horizon
step. The gain \emph{decays} with horizon on all five databases, from $h=1$ to $h=24$, which is the complement of the
residual correlation rising with horizon: members converge on the same drift, so there is
progressively less disagreement to arbitrate. Table~\ref{tab:horizon} gives the values and
Figure~\ref{fig:horizon} plots them.

\begin{table}[htbp]
\centering\small
\caption{Combination gain against the best single model by horizon step (\%).}\label{tab:horizon}
\begin{tabular}{rrrrrrr}
\toprule
horizon & \textsc{UMM} & HiRID & MIMIC-III & MIMIC-IV & eICU & mean \\
\midrule
1 & -20.99 & -14.51 & -15.35 & -9.80 & -11.58 & -14.45 \\
2 & -15.48 & -12.40 & -16.29 & -8.15 & -11.08 & -12.68 \\
3 & -9.19 & -25.65 & -14.63 & -8.07 & -8.20 & -13.15 \\
4 & -8.95 & -10.62 & -11.52 & -10.23 & -9.18 & -10.10 \\
5 & -11.08 & -10.68 & -9.01 & -8.55 & -8.08 & -9.48 \\
6 & -11.13 & -10.42 & -14.31 & -6.47 & -8.14 & -10.09 \\
7 & -9.10 & -18.84 & -11.76 & -7.29 & -6.74 & -10.75 \\
8 & -7.23 & -8.72 & -10.17 & -8.29 & -6.85 & -8.25 \\
9 & -12.14 & -11.77 & -4.45 & -6.82 & -6.09 & -8.25 \\
10 & -10.53 & -10.20 & -11.41 & -5.49 & -6.12 & -8.75 \\
11 & -9.74 & -18.61 & -10.19 & -6.38 & -5.63 & -10.11 \\
12 & -11.63 & -8.00 & -8.97 & -7.12 & -5.42 & -8.23 \\
13 & -12.05 & -9.02 & -4.37 & -7.69 & -4.94 & -7.61 \\
14 & -8.97 & -9.28 & -10.63 & -4.87 & -5.06 & -7.76 \\
15 & -6.98 & -16.19 & -9.10 & -5.79 & -4.82 & -8.58 \\
16 & -10.05 & -7.73 & -7.62 & -6.22 & -4.55 & -7.23 \\
17 & -11.21 & -9.67 & -5.61 & -7.34 & -4.47 & -7.66 \\
18 & -8.13 & -8.13 & -9.89 & -4.92 & -4.09 & -7.03 \\
19 & -7.06 & -13.20 & -8.29 & -5.21 & -4.11 & -7.57 \\
20 & -12.63 & -8.08 & -6.98 & -5.46 & -3.93 & -7.42 \\
21 & -11.53 & -9.28 & -4.44 & -5.48 & -3.71 & -6.89 \\
22 & -8.76 & -8.26 & -8.68 & -3.99 & -3.96 & -6.73 \\
23 & -7.06 & -8.86 & -7.61 & -4.47 & -3.80 & -6.36 \\
24 & -10.31 & -5.48 & -6.30 & -5.27 & -3.27 & -6.13 \\
\bottomrule
\end{tabular}

\end{table}

\begin{figure}[htbp]
\centering
\includegraphics[width=0.72\textwidth]{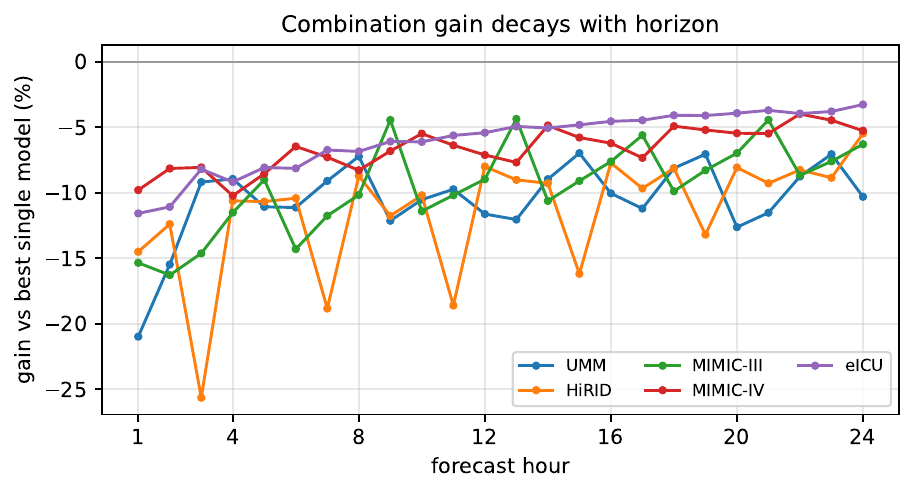}
\caption{Combination gain by horizon. The gain decays as members converge on the same drift. The two smallest databases are visibly noisier hour to hour, the per-hour form of the split-seed spread in Section~\ref{sec:findings}. HiRID's excursion at $h=3$ is a period-4 oscillation present in the underlying data rather than a plotting artifact; it recurs at $h=7$, $11$ and $15$. Its period equals the 4-hour window stride, which is the likely but unverified cause.}
\label{fig:horizon}
\end{figure}

\clearpage

\section{Does the weight profile matter?}\label{app:profile}

Section~\ref{sec:fixed} shows that choosing members per channel is worth about
$2$\,pp while the harmonic profile on those members is worth about $0.2$. This appendix adds the
third rung: instead of fixing the profile at $1,\tfrac12,\dots,\tfrac1k$, \emph{fit} one weight
vector over rank positions, shared by every channel. It is cheap --- permuting the model axis of
$G$ and $b$ into rank order gives the rank-space normal equations with no pass over predictions
--- and it changes little, tracking normalized $1/k$ to within $0.045$ at a correlation of $0.99$.

Table~\ref{tab:profile} weights identical per-channel members three ways, and how much the profile
matters turns out to depend on how many members it spans. At $k=3$ and $k=5$ the three columns
agree  on average to within about $0.2$\,pp, and fitting beats the fixed $1/k$ decay only marginally
(by at most $0.21$\,pp on the development means, and not on the test means). At $k=14$ they separate:
\textsc{equal-14}, the unweighted mean, is $3.1$\,pp worse than \textsc{harm-14} on test and worse
even than \textsc{equal-5}, because flat weights over the whole pool give the weakest members the
same say as the best. It is the decay rather than the fitting that carries the full pool ---
\textsc{harm-14} estimates nothing and still edges out \textsc{fit-14}, $-7.89$ against $-7.85$.

\begin{table}[htbp]
\centering\small\setlength{\tabcolsep}{4pt}
\caption{Weight profiles at fixed membership, as percentage change against the best single model (negative is better). \textsc{equal}, \textsc{harm} and \textsc{fit} use the same per-channel top-$k$ members and differ only in how those members are weighted. \textsc{fit-}$k$ fits one weight vector over rank positions, shared by every channel. \textsc{equal-14} is the unweighted mean over the whole pool. \textsc{ens.} is the per-channel shrunk simplex, shown for scale. Referenced from Section~\ref{sec:fixed}.}\label{tab:profile}
\begin{tabular}{lrrrrrrrrrr}
\toprule
database & \textsc{equal-3} & \textsc{harm-3} & fit-3 & \textsc{equal-5} & \textsc{harm-5} & fit-5 & \textsc{equal-14} & \textsc{harm-14} & fit-14 & \textsc{ens.} \\
\midrule
\multicolumn{11}{l}{\emph{development (half of val $\rightarrow$ held-out half)}} \\
\textsc{UMM} & -7.93 & -7.13 & -7.76 & -8.67 & -8.02 & -8.56 & -8.10 & -10.02 & -10.35 & -10.67 \\
HiRID & -3.05 & -3.17 & -3.18 & -3.74 & -4.01 & -4.07 & -4.50 & -5.62 & -5.14 & -8.00 \\
MIMIC-III & -4.88 & -4.53 & -4.73 & -5.05 & -5.12 & -5.13 & -3.63 & -5.91 & -5.67 & -5.61 \\
MIMIC-IV & -4.37 & -4.41 & -4.53 & -4.08 & -4.70 & -4.72 & -1.20 & -4.87 & -5.14 & -5.57 \\
eICU & -3.79 & -3.87 & -3.95 & -3.75 & -4.12 & -3.92 & -3.14 & -5.06 & -4.64 & -5.00 \\
\addlinespace[2pt]
\quad mean & -4.80 & -4.62 & -4.83 & -5.06 & -5.19 & -5.28 & -4.11 & -6.30 & -6.19 & -6.97 \\
\midrule
\multicolumn{11}{l}{\emph{test (all of val $\rightarrow$ test)}} \\
\textsc{UMM} & -8.58 & -8.10 & -8.44 & -8.98 & -8.79 & -9.04 & -8.01 & -9.95 & -10.13 & -10.95 \\
HiRID & -6.96 & -8.08 & -7.48 & -7.36 & -8.52 & -8.07 & -6.26 & -9.73 & -9.13 & -10.01 \\
MIMIC-III & -7.95 & -7.57 & -7.75 & -8.85 & -8.48 & -8.54 & -5.86 & -8.98 & -9.09 & -9.79 \\
MIMIC-IV & -3.99 & -4.14 & -4.18 & -5.05 & -4.97 & -4.83 & -1.07 & -5.30 & -5.31 & -6.51 \\
eICU & -4.21 & -4.46 & -4.49 & -4.62 & -5.01 & -5.00 & -2.67 & -5.48 & -5.61 & -6.09 \\
\addlinespace[2pt]
\quad mean & -6.34 & -6.47 & -6.47 & -6.97 & -7.15 & -7.10 & -4.77 & -7.89 & -7.85 & -8.67 \\
\bottomrule
\end{tabular}
\end{table}

\clearpage

\section{Inference-time injection and fine-tuning}\label{app:neg}

Five variants of feeding a decoder better values during generation, on two databases and two
iterative decoders, as validation MSE-per-var relative to free-running (positive is worse,
Table~\ref{tab:inject}):

\begin{table}[htbp]\centering\small
\caption{Inference-time injection, as percentage change against free-running generation on validation. The masked rows require the future observation mask and are therefore not deployable.}\label{tab:inject}
\begin{tabular}{lrrrr}
\toprule
 & DenseAR \textsc{UMM} & STraTS \textsc{UMM} & DenseAR HiRID & STraTS HiRID \\
\midrule
free-running (absolute) & 0.4853 & 0.5253 & 0.4344 & 0.4715 \\
ensemble, masked & $-0.3$ & $-9.4$ & $+0.3$ & $-3.4$ \\
ground truth, masked & $+1.1$ & $-29.4$ & $+1.7$ & $-23.5$ \\
ensemble, unmasked & $+1.9$ & $+13.7$ & $+0.3$ & $+29.1$ \\
per-channel best, unmasked & $+3.5$ & $+20.5$ & $+0.6$ & $+38.6$ \\
live hybrid & $+3.6$ & $+29.1$ & $+0.6$ & $+178.3$ \\
\bottomrule
\end{tabular}
\end{table}

\noindent The masked rows inject only where the target is observed, so which channels receive an
injection encodes the future observation mask; since ICU sampling is informative, those rows are not
deployable however large the improvement looks. The unmasked rows need no such information and are
uniformly worse. A decoder trained on its own predictions is already adapted to free-running and
even ground truth costs it 1--2\%; a decoder trained under a teacher-forcing curriculum carries
exposure bias instead, which is why injection appears to help it so much and why that help cannot be
used.

Fine-tuning DenseAR on \textsc{UMM} to consume the combined trajectory improved validation error from 0.4245 to
0.4166, but a matched control --- identical fine-tuning, no injection --- reached 0.4187. Of the
apparent $-1.9\%$, therefore, $-1.37$\,pp comes from training longer with a fresh learning-rate schedule,
which the matched control gets as well, and only $-0.49$\,pp is attributable to injection, while the resulting model requires the
whole pool at inference exactly as post-hoc combination does.

\clearpage

\section{Pool size and the extended pool}\label{app:poolsize}


Table~\ref{tab:bymodel} gives the test error of every member and combiner of the extended
15-member pool, and Table~\ref{tab:winshare15} the share of cells each member wins.

\begin{table*}[htpb]
\centering\small\setlength{\tabcolsep}{4pt}
\caption{Test MSE per variable for every member and combiner of the extended 15-member pool
(the original pool without Moirai and TiRex-2, plus LightGBM, ARIMA and ETS). The last column
is the mean over databases of the \emph{relative} change against LightGBM, not a mean of the
absolute values: absolute MSE differs in scale between databases, so averaging it would weight
\textsc{UMM} most. harm-15 is the harmonic blend of all 15 members. Only channels with at least 100
observations, as in Appendix~\ref{app:system}; Table~\ref{tab:lgbm} uses all charted channels.}
\label{tab:bymodel}
\begin{tabular}{lrrrrrr}
\toprule
 & \textsc{UMM} & HiRID & M-III & M-IV & eICU & mean $\Delta$ \\
\midrule
stacker (shrunk simplex) & 0.515 & 0.418 & 0.397 & 0.410 & 0.390 & $\mathbf{-5.3}$ \\
harm-15 & 0.524 & 0.427 & 0.403 & 0.417 & 0.394 & $-3.8$ \\
harm-5 & 0.523 & 0.423 & 0.402 & 0.417 & 0.395 & $-4.0$ \\
per-variable select & 0.539 & 0.431 & 0.415 & 0.429 & 0.408 & $-1.2$ \\
equal weights & 0.544 & 0.454 & 0.422 & 0.436 & 0.407 & $+0.6$ \\
\midrule
LightGBM & 0.530 & 0.450 & 0.420 & 0.436 & 0.411 & $0.0$ \\
Informer & 0.577 & 0.488 & 0.444 & 0.449 & 0.422 & $+5.7$ \\
DenseAR & 0.596 & 0.487 & 0.447 & 0.443 & 0.418 & $+6.1$ \\
GRU & 0.591 & 0.491 & 0.450 & 0.448 & 0.425 & $+6.8$ \\
Dense DMS & 0.580 & 0.497 & 0.455 & 0.453 & 0.435 & $+7.6$ \\
STraTS & 0.601 & 0.540 & 0.471 & 0.470 & 0.447 & $+12.4$ \\
LSTM & 0.649 & 0.547 & 0.473 & 0.457 & 0.427 & $+13.0$ \\
PatchTST & 0.617 & 0.513 & 0.487 & 0.530 & 0.462 & $+16.1$ \\
DLinear & 0.605 & 0.495 & 0.510 & 0.533 & 0.483 & $+17.1$ \\
Mamba & 0.664 & 0.568 & 0.485 & 0.497 & 0.460 & $+18.6$ \\
Autoformer & 0.661 & 0.559 & 0.536 & 0.554 & 0.506 & $+25.4$ \\
\textsc{locf} & 0.702 & 0.591 & 0.562 & 0.624 & 0.526 & $+33.8$ \\
ARIMA & 0.742 & 0.665 & 0.609 & 0.628 & 0.564 & $+42.8$ \\
ETS & 0.777 & 0.675 & 0.633 & 0.700 & 0.605 & $+51.1$ \\
\textsc{null} & 1.060 & 1.010 & 0.933 & 0.986 & 0.995 & $+123.0$ \\
\bottomrule
\end{tabular}
\end{table*}

\begin{table}[htpb]
\centering\scriptsize\setlength{\tabcolsep}{3.5pt}
\caption{Share of well-supported (variable, hour) test cells won by each member of the extended
15-member pool (\%). This pool excludes Moirai and TiRex-2, so the shares are not comparable
with Table~\ref{tab:winshare}, which covers the original 14-member pool. Columns and the mean
each sum to 100\,\%.}
\label{tab:winshare15}
\begin{tabular}{lrrrrrr}
\toprule
member & \textsc{UMM} & HiRID & M-III & M-IV & eICU & mean \\
\midrule
LightGBM & 75.7 & 74.5 & 61.9 & 56.8 & 58.5 & \textbf{65.5} \\
DenseAR & 1.6 & 1.7 & 13.6 & 6.0 & 16.4 & 7.9 \\
STraTS & 1.4 & 6.4 & 5.4 & 13.3 & 6.5 & 6.6 \\
PatchTST & 3.2 & 2.1 & 5.4 & 4.9 & 5.2 & 4.2 \\
DLinear & 8.1 & 8.5 & 0.8 & 1.2 & 0.7 & 3.8 \\
GRU & 1.3 & 0.9 & 0.8 & 7.1 & 2.3 & 2.5 \\
Informer & 3.1 & 1.2 & 1.8 & 3.6 & 1.1 & 2.2 \\
Dense DMS & 2.2 & 0.5 & 2.9 & 0.9 & 4.1 & 2.1 \\
\textsc{locf} & 0.9 & 1.9 & 3.1 & 1.8 & 2.8 & 2.1 \\
\textsc{null} & 0.1 & 0.0 & 2.8 & 1.8 & 0.1 & 1.0 \\
LSTM & 1.4 & 0.4 & 0.1 & 0.6 & 1.2 & 0.8 \\
ARIMA & 0.3 & 0.7 & 0.6 & 1.2 & 0.1 & 0.6 \\
Autoformer & 0.3 & 0.4 & 0.3 & 0.1 & 0.8 & 0.4 \\
ETS & 0.1 & 0.7 & 0.4 & 0.5 & 0.1 & 0.4 \\
Mamba & 0.3 & 0.0 & 0.0 & 0.1 & 0.2 & 0.1 \\
\bottomrule
\end{tabular}
\end{table}

\begin{table}[htbp]
\centering\small\setlength{\tabcolsep}{4pt}
\caption{Controls behind the Discussion (Section~\ref{sec:discussion}), in percent (negative is
better). First row: 1\,000 random convex weight vectors per database over the original 14-member pool, on test, against
\textsc{best-single}; brackets give the 5th and 95th percentile. Remaining rows, development
protocol: \textsc{ensemble} fitted on four seeds of one architecture, against that architecture's
best seed; and the error of the stacked diverse 14-member pool relative to each seed stack.}
\label{tab:controls}
\begin{tabular}{lrrrrr}
\toprule
 & \textsc{UMM} & HiRID & MIMIC-III & MIMIC-IV & eICU \\
\midrule
random convex weights & $-6.8$ {\tiny [-7.6, -5.9]} & $-4.7$ {\tiny [-6.3, -2.8]} & $-4.4$ {\tiny [-5.3, -3.5]} & $+0.5$ {\tiny [-0.7, +2.1]} & $-1.1$ {\tiny [-1.8, -0.4]} \\
\addlinespace[2pt]
4 DenseAR seeds vs.\ best seed & $-3.9$ & $-4.9$ & $-3.4$ & $-2.6$ & $-1.1$ \\
4 STraTS seeds vs.\ best seed & $-5.1$ & $-2.0$ & $-1.9$ & $-8.2$ & $-6.1$ \\
\addlinespace[2pt]
diverse pool vs.\ DenseAR seeds & $-12.8$ & $-13.6$ & $-4.7$ & $-4.1$ & $-4.2$ \\
diverse pool vs.\ STraTS seeds & $-17.1$ & $-21.8$ & $-8.3$ & $-20.6$ & $-14.6$ \\
\bottomrule
\end{tabular}
\end{table}

\begin{figure}[htpb]
\centering
\begin{tikzpicture}
\begin{axis}[
  width=0.95\linewidth, height=4.1cm,
  xlabel={members in the pool}, ylabel={gain over best single (\%)},
  xtick={2,4,6,8,10,12,14}, ymin=-12, ymax=1,
  grid=major, grid style={gray!20},
  legend style={font=\scriptsize, at={(0.5,1.03)}, anchor=south, legend columns=5, draw=none},
  tick label style={font=\scriptsize}, label style={font=\small},
]
\addplot coordinates {(2,-2.25)(4,-8.20)(6,-10.39)(8,-10.68)(10,-10.66)(12,-10.78)(14,-10.15)};
\addlegendentry{\textsc{UMM}}
\addplot coordinates {(2,-5.64)(4,-9.78)(6,-8.56)(8,-9.32)(10,-8.72)(12,-9.54)(14,-9.92)};
\addlegendentry{HiRID}
\addplot coordinates {(2,-3.33)(4,-4.20)(6,-4.81)(8,-5.32)(10,-5.16)(12,-5.26)(14,-5.54)};
\addlegendentry{MIMIC-III}
\addplot coordinates {(2,-2.94)(4,-3.58)(6,-5.14)(8,-5.52)(10,-5.37)(12,-5.36)(14,-5.57)};
\addlegendentry{MIMIC-IV}
\addplot coordinates {(2,-3.42)(4,-6.12)(6,-4.50)(8,-4.19)(10,-5.00)(12,-4.96)(14,-5.00)};
\addlegendentry{eICU}
\addplot[black, dashed, thick, forget plot] coordinates {(2,0)(14,0)};
\end{axis}
\end{tikzpicture}
\caption{Random subsets of the 14-member pool, five draws per size, scored with the
shrunk-simplex combiner against the full pool's best single model (negative is better).
Each point is the mean over the five draws; the rightmost point at 14 is the full pool, not a
subset. Development protocol: fitted on one stay-grouped half of validation and scored on the
other, so these are not test gains. Four to six members recover about $90\,\%$ of the full
pool's gain and eight recover $97\,\%$. Only 4 of
the 150 subsets (six sizes from 2 to 12, five draws each, five databases) are worse than the best single model, and all four are individual draws at
$k=2$, which the means hide. The subsets are random rather than chosen on validation, so a
guided choice would likely need fewer.}
\label{fig:poolsize}
\end{figure}
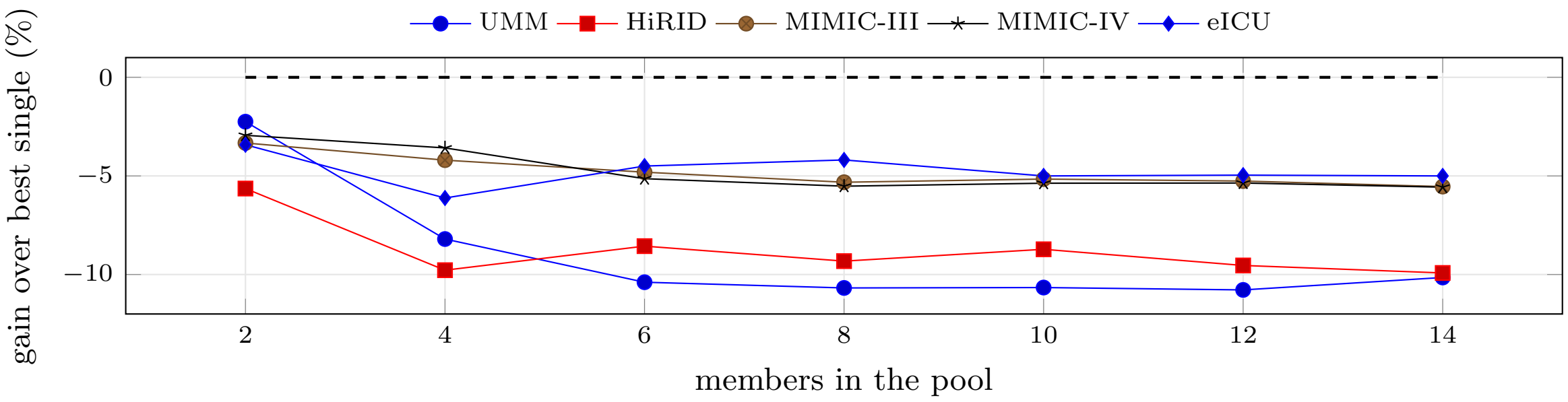

\end{document}